\documentclass[journal]{IEEEtran}
\usepackage{epsfig}
\usepackage{graphicx}
\usepackage{subfigure}
\usepackage[cmex10]{amsmath}
\usepackage{mathrsfs}
\usepackage{amsfonts}
\usepackage{amsmath}
\usepackage{color}
\usepackage{framed}
\usepackage{algorithm}
\usepackage{algpseudocode}

\newtheorem{Problem}{Problem}
\newtheorem{Definition}{Definition}

\DeclareMathOperator{\trace}{tr}

\usepackage{blindtext}
\usepackage{amsbsy}
\usepackage{float}
\usepackage{fancyhdr}
\usepackage{booktabs}
\usepackage{mdwmath}
\usepackage{mdwtab}
\usepackage{eqparbox}
\ifCLASSINFOpdf
\else
\fi
\begin{document}

\title{Simultaneous Spectral-Spatial Feature Selection and Extraction for Hyperspectral Images}

\author{Lefei Zhang,~\IEEEmembership{Member,~IEEE,}
        Qian Zhang,~\IEEEmembership{Member,~IEEE,}
        Bo Du,~\IEEEmembership{Senior Member,~IEEE,}\\
        Xin Huang,~\IEEEmembership{Senior Member,~IEEE,}
        Yuan Yan Tang,~\IEEEmembership{Fellow,~IEEE,}
        and Dacheng Tao, ~\IEEEmembership{Fellow,~IEEE}

\thanks{Manuscript received March 16, 2016, revised July 2, 2016, and accepted August 20, 2016. This paper is supported in part by the National Natural Science Foundation of China under Grants 61401317, 61471274, 91338202 and 41501392, and by the Australian Research Council Projects FT-130101457, DP-140102164 and LE-140100061. (Corresponding author: Bo Du.)}

\thanks{Lefei Zhang and Bu Du are with the State Key Laboratory of Software Engineering, School of Computer, Wuhan University, Wuhan 430072, China.}
\thanks{Qian Zhang is with the Beijing Samsung Telecom R\&D Center, Beijing 100028, China.}
\thanks{Xin Huang is with the School of Remote Sensing and Information Engineering, Wuhan University, Wuhan 430072, China.}
\thanks{Yuan Yan Tang is with the Department of Computer and Information Science, University of Macau, Macau 999078, China.}
\thanks{Dacheng Tao is with Centre for Quantum Computation and Intelligent Systems, Faculty of Engineering and Information Technology, University of Technology, Sydney, NSW 2007, Australia}
\thanks{\copyright20XX IEEE. Personal use of this material is permitted. Permission from IEEE must be obtained for all other uses, in any current or future media, including reprinting/republishing this material for advertising or promotional purposes, creating new collective works, for resale or redistribution to servers or lists, or reuse of any copyrighted component of this work in other works. }
}


\maketitle

\begin{abstract}
    In hyperspectral remote sensing data mining, it is important to take into account of both spectral and spatial information, such as the spectral signature, texture feature and morphological property, to improve the performances, e.g., the image classification accuracy. In a feature representation point of view, a nature approach to handle this situation is to concatenate the spectral and spatial features into a single but high dimensional vector and then apply a certain dimension reduction technique directly on that concatenated vector before feed it into the subsequent classifier. However, multiple features from various domains definitely have different physical meanings and statistical properties, and thus such concatenation hasn't efficiently explore the complementary properties among different features, which should benefit for boost the feature discriminability. Furthermore, it is also difficult to interpret the transformed results of the concatenated vector. Consequently, finding a physically meaningful consensus low dimensional feature representation of original multiple features is still a challenging task. In order to address the these issues, we propose a novel feature learning framework, i.e., the simultaneous spectral-spatial feature selection and extraction algorithm, for hyperspectral images spectral-spatial feature representation and classification. Specifically, the proposed method learns a latent low dimensional subspace by projecting the spectral-spatial feature into a common feature space, where the complementary information has been effectively exploited, and simultaneously, only the most significant original features have been transformed. Encouraging experimental results on three public available hyperspectral remote sensing datasets confirm that our proposed method is effective and efficient.
\end{abstract}

\begin{IEEEkeywords}
    Feature extraction, feature selection, hyperspectral data, spectral-spatial classification
\end{IEEEkeywords}

\IEEEpeerreviewmaketitle

\section{Introduction}

    Over the past two decades, the significant advances in the hypespectral sensors have opened a new way to earth observation in remote sensing \cite{GD2014,LH2012}. These sensors, both space-borne and airborne, simultaneously capture the radiance of materials in hundreds of narrow contiguous spectral bands and result in cube like data. Such data, which is composed of two spatial dimensions (width and height) and a spectral dimension, provides both detailed spectral and structural information for the analysis and recognition of ground materials. Therefore, hyperspectral images have been increasingly applied in many areas including the precision agriculture, military application and environmental management \cite{YL2012,HX2015}. Among these applications, hyperspectral image classification is extremely important and has been attracted by many focuses in recent years \cite{MY2013,YY2016}.

    Previously, most of the multi- and hyper- spectral image classification methods were mainly developed in the spectral domain, based on the idea of the spectral feature contains enough information to infer the label of a pixel \cite{YL2012,WS2012,HY2015}. These feature vectors, which had considered to be fed into the classifier, were represented by the independent spectral characteristics of pixels but without taken into account of the spatial relationship of their neighbor pixels.

    Recently, some investigations had shown the limitations of using only spectral feature and incorporate the spatial information as well to further improve the classification accuracy \cite{YB2014,JY2015,LS2015,HY2016}. In practice, the fact that the adjacent pixels are related or corrected in real images is important for hyperspectral images classification \cite{JH2014,LS2014,HG2014,JY2016,XX2016}. Tarabalka \emph{et al.} \cite{YT2010} proposed a spectral-spatial classification method by marker-based segmentation techniques. However, in \cite{YT2010}, the key was how to select the markers which strongly depended on the result of pixelwise classification. In addition, Huang \emph{et al.} \cite{XL2013} proposed a SVM based multiclassifier model with semantic based postprocessing to ensemble combine spectral and spatial features at both pixel and object levels. Zhong \emph{et al.} \cite{PR2010} formulated a conditional random field to exploit the strong dependencies across spatial and spectral neighbors for hypespectral image classification. All the experimental results had shown that the spectral and spatial methods mentioned above had significantly improved classification accuracy when compared with the previous spectral based techniques.

    However, most of the aforementioned methods which had employed different tactics to incorporate the spatial information, could be regarded to address a particular postprocessing step to boost the subsequent image classification performance. In this paper, we exploit both the spectral and spatial information from the perspective of feature representation, since we believe that the efficient feature representation is the key engine of the subsequent classification task \cite{YA2013,DS2014}. Different feature representations contain more or less different information of the observed data, and the discriminability of feature representation directly decides the upper boundary of classification results \cite{XB2013}. As long as the feature representation contains enough discriminative information what the classification needs, we can get a satisfactory result. Recently, to discover the spectral-spatial feature representation in the remote sensing field, some works have emerged \cite{LL2013a,JX2015,LQ2015,JH2015}. A simple and natural method to handle spectral and spatial features is vector stacking, which concatenates different kinds of features into a long vector, unfortunately, stacking the spectral feature and spatial feature would produce a higher dimensional feature vector \cite{LX2016}. The so called curse of dimensionality \cite{G1968}, would occur when the number of available training samples is limited \cite{JH2015,YG2015}. To deal with this issue, Fauvel \emph{et al.} proposed an approach to fuse morphological information and the original spectral data via reduction of dimensionality \cite{MJ2008}. Additionally, Zhang \emph{et al.} \cite{LL2013b} proposed a multiple features combining approach for classification. The experimental results demonstrated that appropriately concatenate the spectral and spatial features could boost the classification accuracy.

    Nevertheless, the feature stacking strategy still suffers from some problems when apply to hyperspectral image spectral-spatial feature representation and  classifications. Firstly, it treats different features equally and thus ignores the specifical properties of multiple features. Secondly, it fails to explore both the consistent information of different features and the complementary information among multiple features. Last but not least, when some feature extraction techniques (e.g., the Principal Component Analysis, PCA) are directly applied on the stacked vectors, there is a major disadvantage that the learned projection is a linear combination of all the original candidate features. Therefore, it is difficult to interpret which feature in the original feature set plays an essential role in the classification. In view of above problems, in this paper, we study a new spectral-spatial feature learning method for hyperspectral image classification, which termed simultaneous spectral-spatial feature selection and extraction, or S3FSE for short, motivated by the recent advance in manifold learning \cite{TD2009} and structured sparse learning \cite{FH2010}. In particular, our proposed method integrates the feature selection and feature extraction into a joint framework to perform hyperspectral image spectral-spatial feature learning, by which the learned result could be interpretable. In detail, the major contributions of this paper are summarized as follows.

    \begin{itemize}
      \item We propose a novel multiple features learning method by integrating the merits of both feature selection and feature extraction, which could discern the importance of original feature set and alleviate the drawback of feature extraction that the transformed result is difficult to interpret.
      \item The advantage of manifold learning is incorporated into our framework to capture the relation among multiple features. Co-local geometric preserving is proposed to preserve the geometric properties of multiples features. Meanwhile, a co-graph regularization is proposed to exploit both the consistent information of different features and the complementary information among multiple features.
      \item To avoid the problem of simultaneously optimize the selection matrix and transformation matrix, the $\ell_{2,1}$-norm is exerted to co-regularize the different projection matrices to obtain row sparse. By projecting the spectral-spatial feature into a common feature space, the redundant features and noises have been discarded, and only the significant original features have been transformed.
    \end{itemize}

    The remainder of this paper is structured as follows. In Section II we present the objective function of the proposed simultaneous spectral-spatial feature selection and extraction method. In section III, we provide the detailed optimization steps of our proposed method. The experimental results on three public available hyperspectral datasets are reported in Section IV, followed by the conclusions in Section V.

\section{The Proposed Method}

\begin{figure*}[htb]
    \centering
      \includegraphics[width = 5.5 in, keepaspectratio]{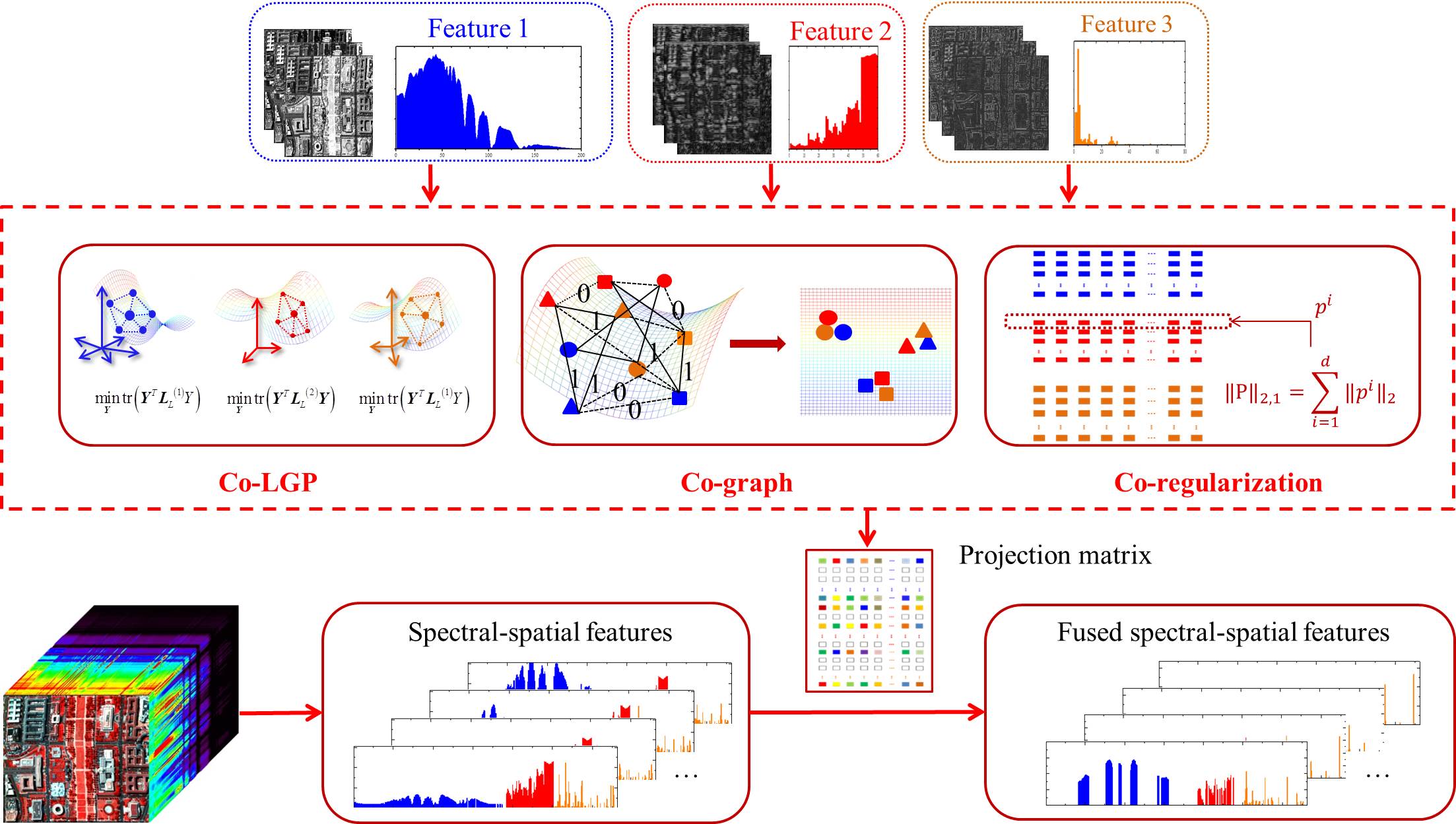}
      \caption{The Flowchart of the proposed approach.}
      \label{fig1}
    \end{figure*}

    In this section, we describe the proposed method for hyperspectral image feature learning in detail. The proposed S3FSE can be divided into three main components, as shown in Fig. \ref{fig1}. In the first step, spectral and spatial features are extracted for each pixel. Then, based on manifold learning and structure sparse learning, the structure information of data is exploited. Finally, a row sparse projection matrix is learned, which is able to discard the redundant and noisy features and transform the significant original features simultaneously.

\subsection{Notations and Problem Definition}
    Before going to the detail of the proposed algorithm, we firstly summarize some notations used throughout this paper. We utilize uppercase letters to denote matrices, and bold lowercase letter to denote vectors. For a matrix $M\in \mathbb{R}^{a\times b}$, its $i$-th row and $j$-th column are denoted as $m^i$ and $m_j$, respectively. $M_{ij}$ means the $(i, j)$-th entry of $M$. And the $\ell_{r,p}$-norm of $M$ is defined as follows.

\begin{equation} \label{formula1}
    \left\Vert M \right\Vert_{r,p} = ({\textstyle \sum_{i=1}^{u} }({\textstyle \sum_{j=1}^{v}\left| M_{ij} \right|}^r)^{\frac{p}{r}})^{\frac{1}{p}}
\end{equation}

    Let $\textbf{q}=\{q_1,q_2,\cdots,q_{\ell}\}$ to be a set of pixels in the hyperspectral image, where $\ell$ is the number of pixels. Assume that the set of samples can be represented in $V$ spectral and spatial feature spaces, or with $V$ heterogeneous feature representations (for convenience, following the perspective of multiview learning \cite{CD2015}, we call each type of feature representation as a view in next paragraphs). Let $\mathcal{F} =\{F^{(1)},F^{(2)},\cdots,F^{(V)}\}$ to be the set such feature spaces and $F^{(i)}\in \mathbb{R}^{d_i}$ be the feature space for the $i$-th view, where $d_i$ denotes the dimensionality of the $i$-th feature space $F^{(i)}$. Denote $D = \{X^{(i)}\in \mathbb{R}^{\ell \times d_i }\}_{i=1}^V$ as the set of views and in which $X^{(i)}$ is the feature matrix of the $i$-th view.

    \begin{Problem}[\textbf{Learning Objective}]
     Given $n$ training samples $D_{training} =\{X^{(i)}\in \mathbb{R}^{n \times d_i}\}_{i=1}^V$ with their labels $\{\textbf{z}_i\}_{i=1}^{n}$ from $C$ classes, our objective is to learn a latent low dimensional feature representation (denoted as $Y \in \mathbb{R}^{n\times d}$, in which $d$ is the  dimensionality of the learned subspace) of the input multiple features, which can achieve a better performance of hyperspectral image classification.
    \end{Problem}

\subsection{The Objective Function of S3FSE}
     In this paper, we propose a simultaneous spectral-spatial feature selection and extraction algorithm (S3FSE) to achieve the aforementioned learning objective. S3FSE learns a shared latent low dimensional subspace by project the spectral and spatial features into a common space. To derive that common low dimensional subspace for spectral and spatial features, two important issues should be taken into account. One is the common low dimensional representation should preserve the local geometric structures in the original spectral and spatial feature spaces, respectively. The other one is that the within-class variation of the common low dimensional representation should be minimized by taking advantage of the consistent information and complementary information of the spectral and spatial features. In S3FSE, we adopt manifold learning and structure sparse learning techniques to obtain the goals above. In the proposed algorithm, the graph Laplacian based on the patch alignment framework \cite{TD2009} is used to characterize the local geometric structures of heterogeneous features. In order to minimize the within class variation, S3FSE characterizes the consensus data description for heterogeneous feature representations by a co-graph regularization. Meanwhile, to make the learned subspace interpretable, S3FSE inherits the advantage of feature selection by a co-regularization of projection matrices. Consequently, there are mainly three parts in the proposed  objection function, which will be introduced in sequence.

     \textbf{Co-local geometric preserving (CoLGP).} Motivated by the intuition that nearby data points have similar geometric properties \cite{DX2011,JD2012,YH2014}, we construct graph Laplacian to model the local neighborhood relationships data points. Let $G^{(v)} = \{X^{(v)},W^{(v)}\}$ to be a undirected weighted graph with vertex set $X^{(v)}$ and weighed matrix $W^{(v)}\in \mathbb{R}^{n\times n}$ for the $v$-th view. Denote $N(\textbf{x}^{(v)}_i)$ as the set of k-nearest-neighbors of $\textbf{x}^{(v)}_i$ by the Euclidean distance metric, then each element $w^{(v)}_{ij}$ of the weighted matrix $W^{(v)}$ is defined as follows.
\begin{equation} \label{formula2}
w^{(v)}_{ij} =
\begin{cases}
e^{(-\frac{\left\Vert \textbf{x}^{(v)}_i -\textbf{x}^{(v)}_j \right\Vert^2 }{ t })},& {\textbf{x}^{(v)}_j \in N(\textbf{x}^{(v)}_i)~or~\textbf{x}^{(v)}_i \in N(\textbf{x}^{(v)}_j)}\\
0,& {otherwise}
\end{cases}
\end{equation}

    Hence, the local geometrical structure of the $v$-th view can be considered by:
\begin{equation} \label{formula3}
\min_{Y} \sum _{i\neq j}{w^{(v)}_{ij}\left\Vert \textbf{y}_i - \textbf{y}_j \right\Vert^2}
\end{equation}
    where $\textbf{y}_i$ and $\textbf{y}_j$ are the shared low dimensional representation for samples $i$ and $j$, respectively.

    Based on the Laplacian Eigenmaps \cite{XP2004}, Eq. \eqref{formula3} can be reformulated to:
\begin{equation} \label{formula4}
\min_{Y} \; \trace(Y^{\rm{T}}L^{(v)}Y), ~s.t.\quad Y^{\rm{T}}Y = I
\end{equation}
in which $L^{(v)} = D^{(v)} - W^{(v)}$ is the Laplacian matrix, and $D^{(v)}$ is a diagonal matrix whose entries are column sums of $W^{(v)}$.

    Therefore, for heterogenous features from all the $V$ views, we can obtain the following objective function of co-local geometric preserving:
\begin{equation} \label{formula5}
\min_{Y} \; \sum_{v=1}^V\trace(Y^{\rm{T}}L^{(v)}Y)~s.t.  \quad Y^{\rm{T}}Y = I
\end{equation}

    Due to the complementary property provide by heterogenous features, the spectral and spatial features definitely have different contributions to the shared latent low dimensional subspace learning. In order to explore the different contributions of different features, we impose a set of nonnegative weights $\lambda = \{\lambda_1,\lambda_2,\cdots,\lambda_V\}$ on Eq. \eqref{formula5} to better preserve the local geometric properties of different features and explore the complementary property of multiple views at the same time \cite{TD2010,YD2015}, therefore, Eq. \eqref{formula5} can be further written as following:
\begin{equation} \label{formula6}
\min_{Y} \; \sum_{v=1}^V \lambda_v\trace(Y^{\rm{T}}L^{(v)}Y)~s.t.  \quad Y^{\rm{T}}Y = I
\end{equation}

    Let $P = [P^{(1)},P^{(2)},\cdots,P^{(V)}]^{\rm T} \in \mathbb{R}^{ \sum_{v=1}^{V} d_v \times d}$ to be the projection matrix, where $P^{(v)} \in \mathbb{R}^{d_v \times d}$ is the projection matrix of $v$-th view. Considering the different contributions of heterogeneous features, we have:
\begin{equation} \label{formula7}
Y = \sum_{v=1}^V \mu_v X^{(v)}P^{(v)}
\end{equation}
    where $\mu_1,\mu_2,\cdots,\mu_V > 0$. Obviously, $\mu_v = \sqrt{\lambda_v}$. Then, by combing Eqs. \eqref{formula6} and \eqref{formula7}, we can reformulate Eq. \eqref{formula6} to the following form:
\begin{equation} \label{formula8}
\begin{split}
\min_{P} \; &\sum_{v=1}^V \lambda_v\trace({P^{(v)}}^{\rm{T}}{X^{(v)}}^{\rm{T}}L^{(v)}X^{(v)}P^{(v)})\\
=& \trace(\overline{P}^{\rm{T}}H_1\overline{P}) \\
s.t. &\quad \overline{P}^{\rm{T}}X^{\rm{T}}X\overline{P} = I
\end{split}
\end{equation}
    where
\begin{equation} \label{formula9}
\overline{P} = [\sqrt{\lambda_1}P^{(1)},\sqrt{\lambda_2}P^{(2)},\cdots,\sqrt{\lambda_V}P^{(V)}]^{\rm{T}}
\end{equation}
    and
\begin{equation} \label{formula10}
H_1 = \begin{pmatrix}  {X^{(1)}}^{\rm {T}}L^{(1)}X^{(1)} & & \\  & \ddots & \\  &  &  {X^{(V)}}^{\rm {T}}L^{(V)}X^{(V)} \end{pmatrix}
\end{equation}

    As a result, the first part of the objective function is:
\begin{equation} \label{formula11}
\arg \min_{\overline{P}}   \trace(\overline{P}^{\rm{T}}H_1\overline{P})~s.t. \quad \overline{P}^{\rm{T}}X^{\rm{T}}X\overline{P} = I
\end{equation}

    \textbf{Co-graph regularization.} Following the perspective of the multiview learning, on the one hand, heterogeneous features should have different strengths to explore the intrinsic data structure since they have provided complementary information among each other. On the other hand, features from different views should also supply the consistency information by sharing the same semantic label space \cite{CD2014}. In the proposed S3FSE, the problem of exploring the provided complementary information and consistency information from spectral and spatial features can be interpreted to seek a consensus data representation in the shared low dimensional subspace, where the variation of within-class is minimized while the variation of between class is maximized.

\begin{Definition}
Give a set of data representation from heterogeneous features $D = \{X^{(v)} \in \mathbb{R}^{n \times d_v}\}^{V}_{v=1}$ and the projection matrix to the shared common low dimensional subspace $\overline{P} = \{\overline{P}^{(v)}\}_{v=1}^{V}$. Denote $X^{(v)} =[\textbf{x}_1^{(v)},\textbf{x}_2^{(v)},\cdots,\textbf{x}_n^{(v)}]^{\rm{T}} \in \mathbb{R}^{n\times d_v}$ as the matrix representation of $v$-th view. Then,  the Euclidean distance in the low dimensional subspace between data points $i$ and $j$ from view $s$ and $t$ is defined as follows \cite{XY2013}:
\begin{equation} \label{formula12}
d^2(\textbf{x}_i^{(s)},\textbf{x}_j^{(t)}) = \left\Vert {\overline{P}^{(s)}}^{\rm{T}}\textbf{x}_i^{(s)}-{\overline{P}^{(t)}}^{\rm{T}}\textbf{x}_j^{(t)} \right\Vert ^2_2
\end{equation}
\end{Definition}

    For the consensus data description, correspondence pairs in the common low dimensional subspace should be as near as possible. That is to say, the distance between  within-class data points should be as small as possible. In light of Eq. \eqref{formula12}, all views of data points have been embedded into the common low dimensional subspace. Denote $\mathcal{O} = [X^{(1)}\overline{P}^{(1)},X^{(2)}\overline{P}^{(2)},\cdots,X^{(V)}\overline{P}^{(V)}]^{\rm{T}} \in \mathbb{R}^{nV \times d}, (nV = \sum_{v=1}^V n)$ as all views of samples in the common low dimensional subspace, we construct a joint Laplacian graph on the $\mathcal{O}$ to explore the consensus data description \cite{XY2013}. Let $G = \{\mathcal{O},W\}$ to be a joint undirected weighted graph with vertex set $\mathcal{O}$ and weighed matrix $W\in \mathbb{R}^{nV\times nV}$, in which $w_{ij}$ measures the similarity between the data points $i$ and $j$ on $\mathcal{O}$. According to the label information, $W$ is defined as:
\begin{equation} \label{formula13}
w_{ij} =
\begin{cases}
1,&{(i,j)\in c, i\neq j}\\
0,& {otherwise}
\end{cases}
\end{equation}
    where $c$ indicates the $c$-th class. Consequently, the consensus data description in the low dimensional subspace can be exploited by:
\begin{equation} \label{formula14}
\min_{\mathcal{O}} \sum^{nV}_{i\neq j} w_{ij} \left\Vert \mathcal{O}_{i,:} -\mathcal{O}_{j,:} \right\Vert ^2
\end{equation}

Similar to Eq. \eqref{formula4} above, Eq. \eqref{formula14} can be reduced to:
\begin{equation} \label{formula15}
\min_{\mathcal{O}} \trace(\mathcal{O}^{\rm{T}}L\mathcal{O})
\end{equation}
    where $L$ is defined as $L = D - W$ and $D$ is a $nV\times nV$ diagonal matrix with $d_{ii} = \sum_{j=1}^{nV} w_{ij}$.

    If we further denote:
\begin{equation} \label{formula16}
L = \begin{pmatrix} L_{11} & \cdots & L_{1V}\\ \vdots & \ddots & \vdots\\ L_{V1} & \cdots & L_{VV} \end{pmatrix}
\end{equation}

    Then Eq. \eqref{formula15} can be rewritten as:
\begin{equation} \label{formula17}
\begin{split}
 &\min_{\overline{P}} \sum_{s=1}^V \sum_{t=1}^V\trace({\overline{P}^{(s)}}^{\rm {T}}{X^{(s)}}^{\rm{T}}L^{st}X^{(t)}\overline{P}^{(t)})\\
 &= \trace({\overline{P}}^{\rm {T}}H_2\overline{P})
 \end{split}
\end{equation}
    where
\begin{equation} \label{formula18}
H_2 = \begin{pmatrix} {X^{(1)}}^{\rm{T}}L^{(11)}X^{(1)} & \cdots &{X^{(1)}}^{\rm{T}}L^{(1V)}X^{(V)} \\ \vdots & \ddots & \vdots\\ {X^{(V)}}^{\rm{T}}L^{(V1)}X^{(1)}  & \cdots & {X^{(V)}}^{\rm{T}}L^{(VV)}X^{(V)}  \end{pmatrix}
\end{equation}

    Therefore, we have the following second term of the proposed S3FSE algorithm:
\begin{equation} \label{formula19}
\arg \min_{\overline{P}} \trace({\overline{P}}^{\rm {T}}H_2\overline{P})
\end{equation}

    \textbf{Projection matrices co-regularization.} In practice, not all features are important and useful for our classification task, since some of them may be redundant and even noisy. However, one major drawback of the existed feature extraction methods is that the learned projection is a linear combination of all the available features. More often, it is uneasy to interpret the transformed result. Motivated by the characteristic of feature selection which selects a subset of most representative features from the candidate set, in S3FSE, we expect the significant features are transformed by the non-zero values of the projection matrix while the less important features are transformed by zeros value of the projection matrix during the proposed feature extraction. As a result, the transformed result can be regarded as a linear combination of only a subset of all original features. To achieve this goal, one intuitive way is to simultaneously perform both feature selection and feature extraction. The projection matrices can be expressed as following:
\begin{equation} \label{formula20}
\overline{P}^{(v)} = S^{(v)}\odot M^{(v)}
\end{equation}
    where $\odot$ is a Hadamard product operator of matrices. $S^{(v)} \in \mathbb{R}^{d_v\times d}$ is  the selection matrix of feature selection and $M^{(v)}\in \mathbb{R}^{d_v\times d}$ is the transformation matrix of feature extraction for $v$-th view.

    In Eq. \eqref{formula20}, the selection matrix $S^{(v)}$ is defined as:
\begin{equation} \label{formula21}
 S^{(v)}(i,:) =
 \begin{cases}
1,&{feature\;i \; is \;selected}\\
0,& {otherwise}
\end{cases}
\end{equation}

    However, it is not easy to directly solve $S^{(v)}$ and $M^{(v)}$. By combining Eqs. \eqref{formula20} and \eqref{formula21}, we can derive that projection matrix $\overline{P}^{(v)}$ is row sparse. To avoid this problem, we propose to obtain a row sparse projection matrix $\overline{P}^{(v)}$ directly rather than solve the $S^{(v)}$ and $M^{(v)}$.  As indicated in \cite{FH2010,HL2016}, $\ell_{2,1}$ norm measures the distance in feature dimensions and perform summation over different data points via $\ell_2$ and $\ell_1$, respectively. When minimize the $\ell_{2,1}$ norm of the projection matrix, some rows of the matrix would shrink to zeros, thus the $\ell_{2,1}$ norm leads to row sparse as well as exploits the correlations between different features. We therefore resort to $\ell_{2,1}$ norm regularization of projection matrix $\overline{P}^{(v)}$ to make $\overline{P}^{(v)}$ row sparse. The regularization on the projection matrices is then given by:
\begin{equation} \label{formula22}
\Omega(\overline{P}^{(1)},\overline{P}^{(2)},\cdots,\overline{P}^{(V)}) = \min \sum_{v=1}^V \left\Vert \overline{P}^{(v)}\right\Vert_{2,1}
\end{equation}

    It should be noted that Eq. \eqref{formula22} ignores the complementary information from heterogeneous features. In order to address the above mentioned strengths of the multiview data, we utilize $\ell_{2,1}$ norm to co-regulate the projection matrix $\overline{P}$. The co-regularization achieves twofold roles in making the projection matrix $\overline{P}$ row sparse and taking advantages of the complementary information from multiple views. Consequently, the co-regularization of the projection matrices is written as:
\begin{equation} \label{formula23}
\Omega(\overline{P}) = \min_{\overline{P}} \left\Vert \overline{P} \right\Vert_{2,1}
\end{equation}

    Finally, we have the overall objection function of our proposed S3FSE algorithm by integrating Eqs. \eqref{formula11}, \eqref{formula17} and \eqref{formula23} together:
\begin{equation} \label{formula24}
\begin{split}
&\min_{\overline{P}} \; \trace(\overline{P}^{\rm{T}}H_1\overline{P}) + \alpha \trace({\overline{P}}^{\rm {T}}H_2\overline{P}) + \beta \left\Vert \overline{P} \right\Vert_{2,1}\\
&s.t. \quad \overline{P}^{\rm{T}}X^{\rm{T}}X\overline{P} = I
\end{split}
\end{equation}
    where the matrices $H_1$ and $H_2$ have been defined in Eqs. \eqref{formula10} and \eqref{formula18}, and $\alpha$ and $\beta$ are two regularization parameters, respectively.

\section{Optimization}
    Obviously, our objective function Eq. \eqref{formula24} is convex, since $H_1$ and $H_2$ are both positive semi-definite. It can obtain the global optimal solution. However, the objective function in Eq. \eqref{formula24} contains a non-smooth regularization term, i.e., the $\ell_{2,1}$-norm, and in general, it can't be easily solved.

    Denote $\overline{P} = [\overline{\textbf{p}}^1,\overline{\textbf{p}}^2,\cdots,\overline{\textbf{p}}^m]^{\rm{T}} \; (m= \sum_{v=1}^{V} d_v)$ with $\overline{\textbf{p}}^i$ as its $i$-th row. Following \cite{FH2010}, we may reformulate Eq. \eqref{formula24} as:
\begin{equation} \label{formula25}
\begin{split}
&\min_{\overline{P}} \; \trace(\overline{P}^{\rm{T}}H_1\overline{P}) + \alpha \trace({\overline{P}}^{\rm {T}}H_2\overline{P}) + \beta \trace({\overline{P}}^{\rm {T}}H_3\overline{P})\\
&s.t. \quad \overline{P}^{\rm{T}}X^{\rm{T}}X\overline{P} = I
\end{split}
\end{equation}
    where $H_3$ is a diagonal matrix defined as:
\begin{equation} \label{formula26}
H_3 = \begin{pmatrix} \frac{1}{2\left\Vert
\overline{\textbf{p}}^1 \right\Vert_2}& & \\ & \ddots & \\& & \frac{1}{2\left\Vert \overline{\textbf{p}}^m \right\Vert_2} \end{pmatrix}
\end{equation}

    Then, we can rewrite our objective function as:
\begin{equation} \label{formula27}
\begin{split}
&\min_{\overline{P}} \; \trace[\overline{P}^{\rm{T}}(H_1 +\alpha H_2 + \beta H_3)\overline{P}]\\
&s.t. \quad \overline{P}^{\rm{T}}X^{\rm{T}}X\overline{P} = I
\end{split}
\end{equation}

    In Eq. \eqref{formula27}, $\overline{P}$ can be obtained by solving the following eigenvalue problem:
\begin{equation} \label{formula28}
(H_1+ \alpha H_2 + \beta H_3)\overline{\textbf{p}} = \eta X^{\rm{T}}X \overline{\textbf{p}}
\end{equation}

    Let $\overline{\textbf{p}}_1,\overline{\textbf{p}}_2,\cdots,\overline{\textbf{p}}_d$ to be the eigenvectors of Eq. \eqref{formula28} corresponding to the $d$ smallest eigenvalues ordered according to $\eta_1\leq \eta_2 \leq \cdots \leq \eta_d $. We therefore obtain the projection matrix $\overline {P} = [\overline{\textbf{p}}_1,\overline{\textbf{p}}_2,\cdots,\overline{\textbf{p}}_d]$ of our proposed method, where $\overline{\textbf{p}}_i =  [{\overline{p}}_{1i} \dots {\overline{p}}_{Vi}]^{\rm{T}} $.

    In view of the above mathematical deduction, $H_3$ is depend on $\overline{P}$, which is exactly the unknown variable we want to optimize. In this paper, we adopt an iterative approach to optimize Eq. \eqref{formula27}, the complete learning procedure is discussed in Algorithm \ref{alg:S3FSE}. Note that it could be theoretically demonstrated that such an alternating optimization procedure rigorously converges to a global optimum \cite{FH2010}.

\begin{algorithm}[htb]
\caption{Simultaneous Spectral-Spatial Feature Selection and Extraction (S3FSE)}
\label{alg:S3FSE}
\begin{algorithmic}[1]
\Require
 Input data $\{X^{(i)}\in \mathbb{R}^{n \times d_i}\}_{i=1}^V$ and their labels $\textbf{z}_i \in {[1,2,\cdots,C]} $;
 regularization parameters $\alpha$ and $\beta$; the target dimensionality $d$ of low dimensional subspace;
\State The iteration set $t=0$; initialize $\overline{P} \in \mathbb{R}^{m\times d}$ randomly;
\State Compute $H_1$ according to Eq. \eqref{formula10};
\State Compute $H_2$ according to Eq. \eqref{formula18};
\Repeat
\State Compute the diagonal matrix $H_3$ according to Eq. \eqref{formula26};
\State Solve the eigenvalue problem defined in Eq. \eqref{formula28};
\State Obtain the $\overline {P}= [\overline{\textbf{p}}_1,\overline{\textbf{p}}_2,\cdots,\overline{\textbf{p}}_d]$;
\State $t = t+1$;
\Until{Convergence}
\Ensure
 Projection matrix $\overline{P} = [\overline{P}^{(1)},\overline{P}^{(2)},\cdots,\overline{P}^{(V)}]^{\rm{T}}$.
\end{algorithmic}
\end{algorithm}

\section{Experiments}
    In this section, we conduct experiments on three public hyperspectral datasets to show the performance of our proposed algorithm. Following the previous feature learning works \cite{JX2015}, we evaluate the performance of our proposed feature representation method in term of classification. In our experiments, we firstly project the spectral and spatial features into the learned low dimensional subspace, and then use the SVM classifier \cite{CC2011} to classify the test samples in that common feature subspace.

\subsection{Datasets description}
    The first dataset is the HYDICE urban hypersepctral image, which is an urban area captured by the hyperspectral digital imagery collection experiment (HYDICE) airborne remote sensing sensor, at the location of Copperas Cove, near Fort Hood, Texas, USA. It is composed of $187$ spectral channels after removed the low SNR bands, and the whole dataset has the size of $307\times307$ pixels.

    The second dataset is the Washington DC dataset, which was also acquired by the HYDICE sensor over a Mall in the Washington DC, USA. The spatial size of this dataset is $1280 \times 307$ pixels, and there are $191$ spectral channels available for our experiment after deleted the water absorption bands.

    The third dataset is ROSIS Pavia city dataset, which was collected by the Reflective Optics System Imaging Spectrometer (ROSIS) at the city of Pavia, Italy. This dataset is consisted of $1400 \times 512$ pixels.  Due to noise, some channels were removed and the remaining 102 spectral channels are used for our experiment.

\subsection{Experiment setup}

     \textbf{The input spectral and spatial features.} In our experiments, we use three kinds of features, i.e., the spectral feature, texture feature, and morphological feature, as a case study to evaluate the performance of our proposed multifeature learning method. For each pixel in the hyperspectral image, each type of feature has been represented as a single feature vector, respectively.

 \begin{enumerate}
 \item \emph{Spectral feature.}  Denote $\nu_i$ as  the reflectance value of the $i$-th spectral channel, the spectral feature of a pixel in the hyperspectral image can be simply represented by reflectance values of its all $l$  channels, i.e., $\textbf{v}_{Spctral} \in \mathbb{R}^l$.

 \item \emph{Texture feature.}  The $2$-D Gabor wavelet is employed to extract texture information \cite{LL2012}. The PCA transformation is applied to extract the first principal component of the hyperspectral image and then the Gabor function is employed to convolute it. In the 2-D Gabor function, we set the scale parameter as $s=0,1,\cdots,4$ and direction parameter as  $d=0,1,\cdots,11$, and therefore derive the texture feature of a pixel $\textbf{v}_{Texture} \in \mathbb{R}^{60}$.

 \item \emph{Morphological feature.} The Differential Morphological Profile (DMP) is employed to describe the other spatial feature in our experiment. The DMP is based on two commonly used morphological operators (opening and closing) to gather the structural information \cite{JJ2005}. Since the DMP is designed for the gray-level images, in our experiments, we also use PCA to find the first $10$ PC images of the hyperspectral data. After that, circular structural elements with R $=$ $2$, $4$, $6$ and $8$ have been used to compute the DMP feature vector, which result in $\textbf{v}_{DMP} \in \mathbb{R}^{80}$.
 \end{enumerate}

    Nevertheless, it must be pointed out that our proposed feature learning method is actually a general framework which is suitable for hyperspectral image classification with any kinds of features as input.

    \textbf{Comparison schemes.} To validate the effectiveness of our proposed algorithm, we compare it with a baseline and several state-of-the-art dimensionality reduction methods, all of which have accepted the same spectral and spatial feature vectors as input. For the baseline, all the original spectral and spatial features are concatenated into a feature vector. As regards to other comparison methods, the state-of-the-art dimensionality reduction methods are performed on concatenated spectral and spatial feature vectors. In addition, two multiview feature learning algorithms have also been addressed, i.e., the co-local geometric preserving (the first term of the proposed S3FSE algorithm) and the multiple features combining \cite{LL2012}. The detailed comparison methods are enumerated as follows.

\begin{itemize}
\item  Baseline
\item  Sparse Principal Component Analysis (SPCA) \cite{HT2006}
\item  Sparse Discriminant Analysis (SDA) \cite{LT2011}
\item  Cosine-based Nonparametric Feature Extraction (CNFE) \cite{JP2010}
\item  Double Nearest Proportion Feature Extraction (DNP)\cite{HB2010}
\item  Co-Local Geometric Preserving (CoLGP)
\item  Multiple Features Combining (MFC) \cite{LL2012}
\end{itemize}

    \textbf{Implementation details.} For the comparison algorithms, the codes of SPCA and SDA are provided by the open source Matlab Toolbox, i.e., SpaSM \footnote{http://www2.imm.dtu.dk/projects/spasm/}, and the parameters of them are set according to their references \cite{HT2006} and \cite{LT2011}, respectively. In CNFE, there are four parameters, i.e., number of nearest neighbors for local mean ($k$), weighting exponents ($r_1$ and $r_2$) and regularization parameter ($\mu$). In our experiments, we set them as $k=5$, $r_1 =2$, $r_2=1$ and $\mu = 0.75$ according to \cite{JP2010}. In DNP, there are three parameters: self-class nearest proportion ($P_s$), the other class nearest proportion ($P_o$) and regularization parameter ($\mu$) and we set them as $1/8$, $1/8$ and $0.75$, respectively \cite{HB2010}. As respect to the shared model parameters in CoLGP, MFC and S3FSE, we empirically set the neighbor size $k=5$ and the kernel width $t=1$. In addition, the trade-off parameters of proposed S3FSE ($\alpha$ and $\beta$) are tuned by using cross validation with the range of $\{10^{-5},10^{-4},\cdots,10^{4}\}$. For subsequent classification task, the multi-class ones versus one support vector machine classifier with Gaussian radial basis function (RBF) kernel is adopted. In detail, we utilize the LibSVM library \footnote{http://www.csie.ntu.edu.tw/cjlin/libsvm/} as the software tool. And the parameters $C$ and $\gamma$ of SVM classifier are decided via a strategy of cross validation \cite{CC2011} within the range of $\{1, 10, 50,100\}$ and $\{0.1, 1.0, 10, 100\}$, respectively.

\subsection{Experiment 1:  HYDICE urban dataset}

\begin{figure}[htbp]
\begin{center}
\subfigure[]{\label{fig2a}
\epsfig{file = 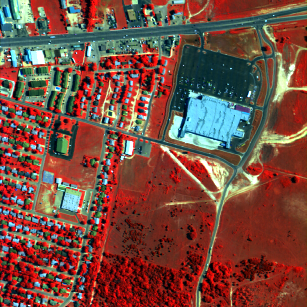, height = 1.2 in, keepaspectratio}}
\subfigure[]{\label{fig2b}
\epsfig{file = 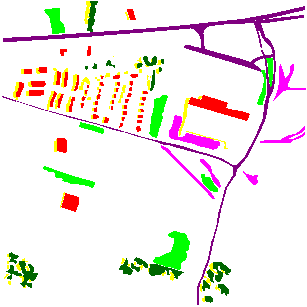, height = 1.2 in, keepaspectratio}}
\subfigure[]{\label{fig2c}
\epsfig{file = 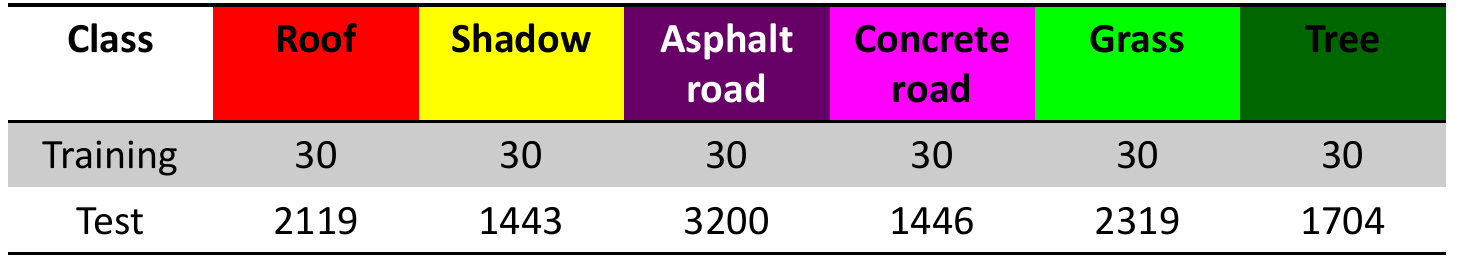, width = 2.8 in, keepaspectratio}}
\caption{ (a) The HYDICE urban dataset, (b) the ground truth map, and (c) the number of training and test samples for classification.}
\label{fig2}
\end{center}
\end{figure}

    In this experiment, we utilize HYDICE urban dataset to detailedly evaluate the performance of our proposed method.  This dataset and its reference data are shown in Figs. \ref{fig2a} and \ref{fig2b}, respectively. According to the ground truth, there are six informative classes of land covers to be analyzed: roof, shadow, asphalt road, concrete road, grass and tree. Although this dataset has a high spatial resolution (around 2m per pixel), yet it is still a challenging one for accurate classification. As we can learn from the previous researches that spectral curves between classes, e.g., roof and road, grass and tree are highly similar \cite{LL2012,HW2014}. Despite from that, we could introduce the spatial features to relieve the misclassifications which lead by only using the spectral feature. As indicated in \cite{LL2012}, although the different land covers may have high correlation coefficients among each other in the spectral domain, we might still distinguish them because they have low correlation coefficients in the spatial feature spaces, e.g., the texture and shape features. Therefore, it is definite that such complementary properties of the multiple features in the hyperspectral image have provided the sufficient information to potentially improve the classification accuracy. In our experiments, in order to verify the effectiveness of our proposed method when very few training samples are available, 30 samples of each class are randomly chosen as the training set. The number of training and test samples is listed in Fig. \ref{fig2c}.

    \textbf{Analysis of the learned projection matrix.} In the spectral and spatial classification, different features have different contributions to explore the essential data structure via providing complementary information. Besides, for the classification task, not all the original features are useful. However, it is usually difficult to interpret the results of the traditional feature extraction method, such as the CNFE and DNP. Although the SPCA and SDA attempt to solve this problem via $\ell_1$-norm regularization, the selected features by sparse methods are independent for each feature dimension. Thus, the results are still difficult to interpret. Our proposed method alleviates this problem via simultaneous performing feature selection and feature extraction. Here we study the learned projection matrix $\overline{P} = [\overline{P}^{(Spectral)}, \overline{P}^{(Texture)}, \overline{P}^{(DMP)}]^{\rm{T}}$ from Eq. \eqref{formula24} with some details. We firstly examine the sparsity of the learned projection matrix. For the HYDICE urban dataset, the overall sparsity of projection matrix $\overline{P}$ is $41.59\%$. In contrast, for spectral feature, texture feature and DMP feature, the sparisties are $22.99\%$, $70.00\%$ and $63.75\%$, respectively. It clearly shows that the spectral feature plays the most important role in the output feature representation. This observation demonstrates that our proposed method effectively explores the complementary characteristics provided by the spectral and spatial features. In addition, the learned projection matrix is row sparse. The non zero rows indicate that the corresponding features are chosen as important features for feature mapping, while the zero rows indicate that the corresponding features are less significant for classification or even be the noise, which should be discarded. Consequently, the low dimensional subspace can be interpreted as a projection from only the significant or relevant subset of original features.

\begin{figure*}[htbp]
\begin{center}
\subfigure[]{\label{fig3a}
\epsfig{file = 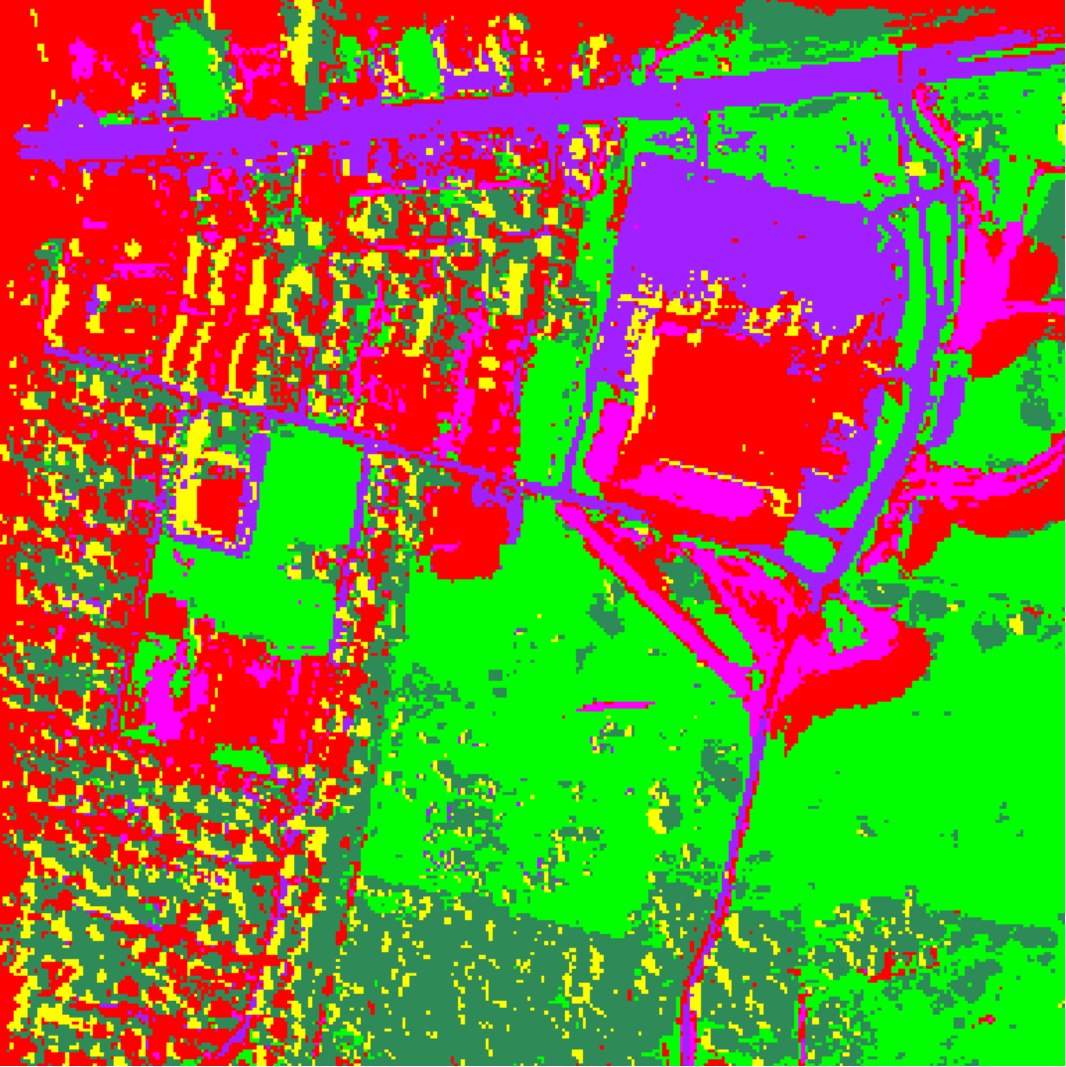, height = 1.2 in, keepaspectratio}}
\subfigure[]{\label{fig3b}
\epsfig{file = 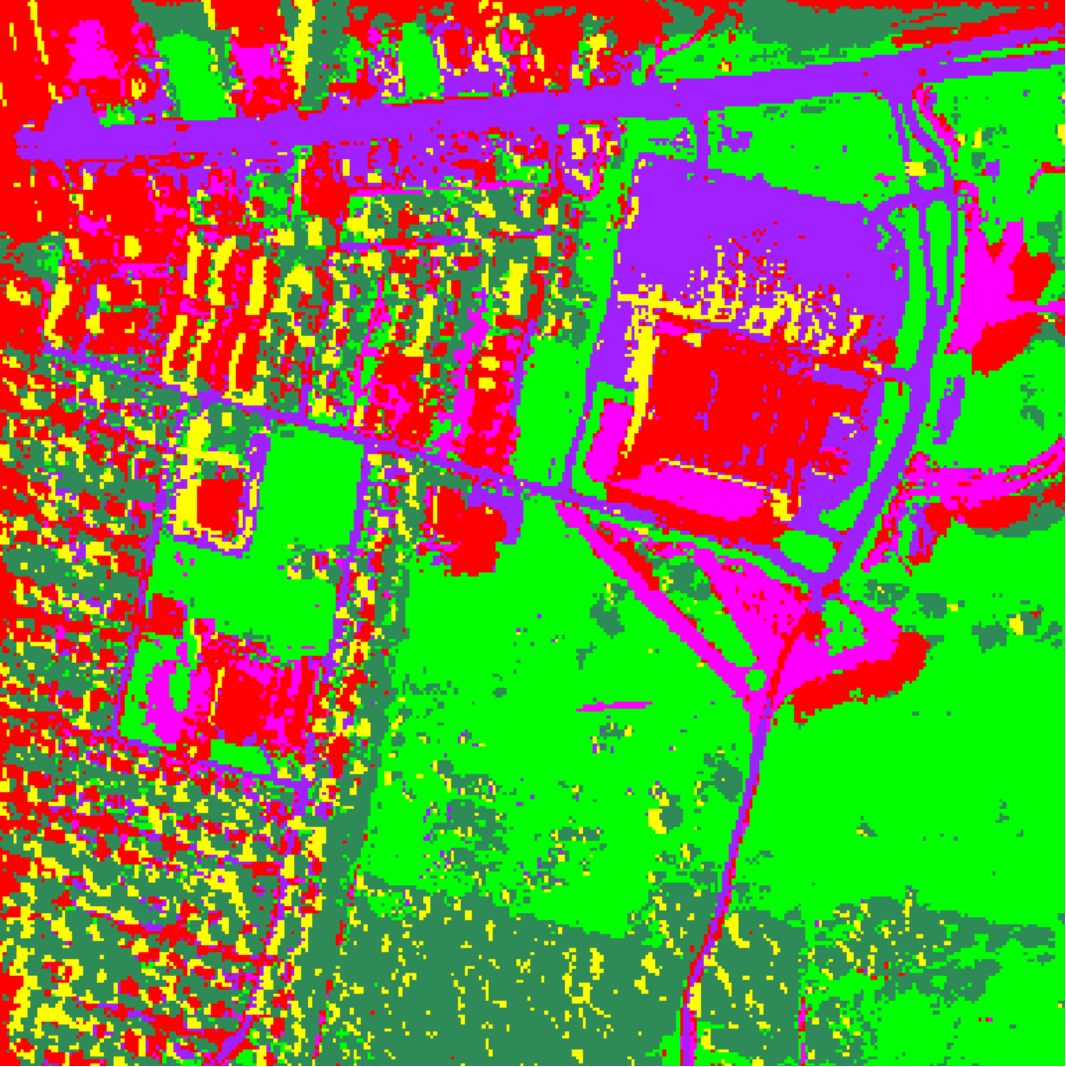, height = 1.2 in, keepaspectratio}}
\subfigure[]{\label{fig3c}
\epsfig{file = 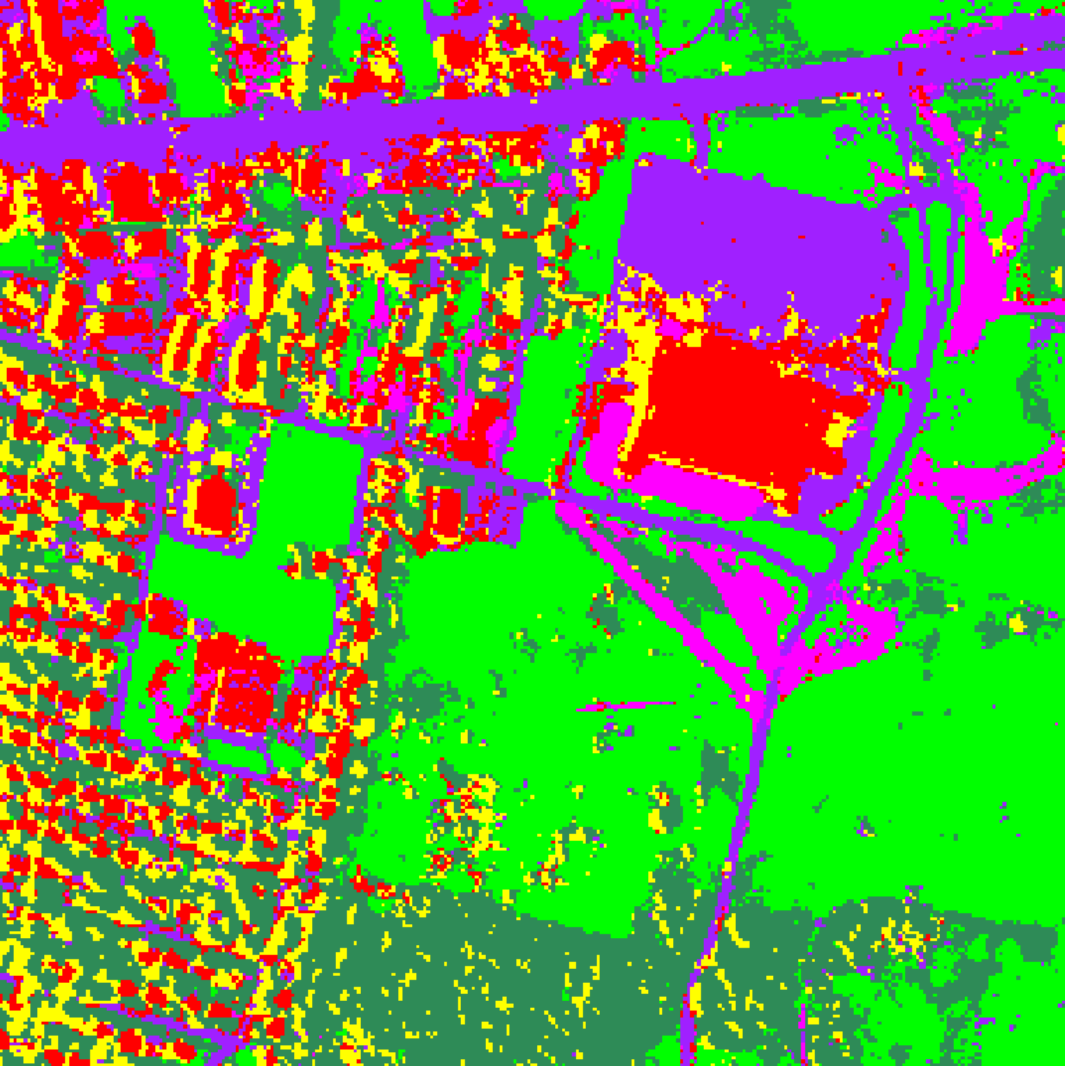, height = 1.2 in, keepaspectratio}}
\subfigure[]{\label{fig3d}
\epsfig{file = 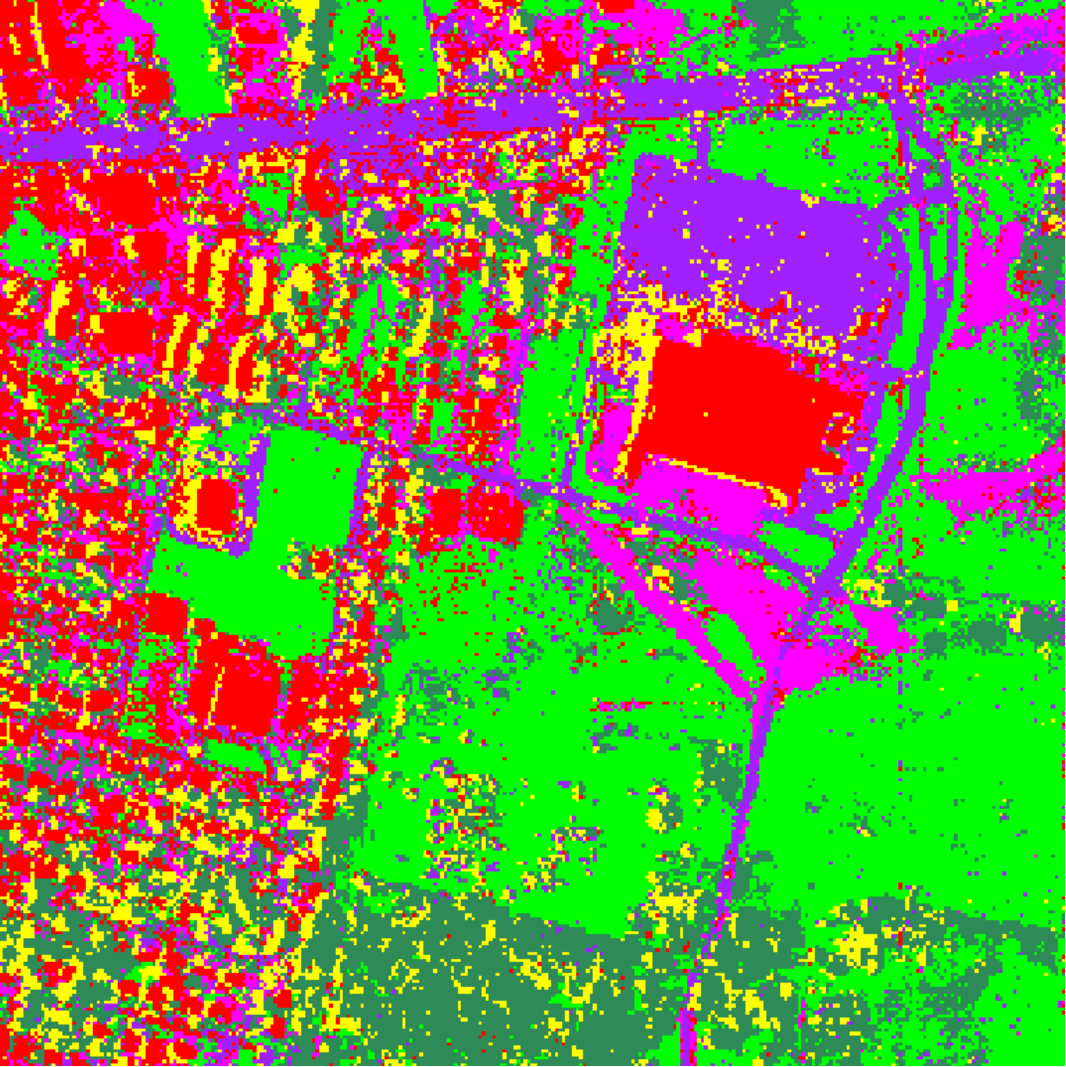, height = 1.2 in, keepaspectratio}}
\subfigure[]{\label{fig3e}
\epsfig{file = 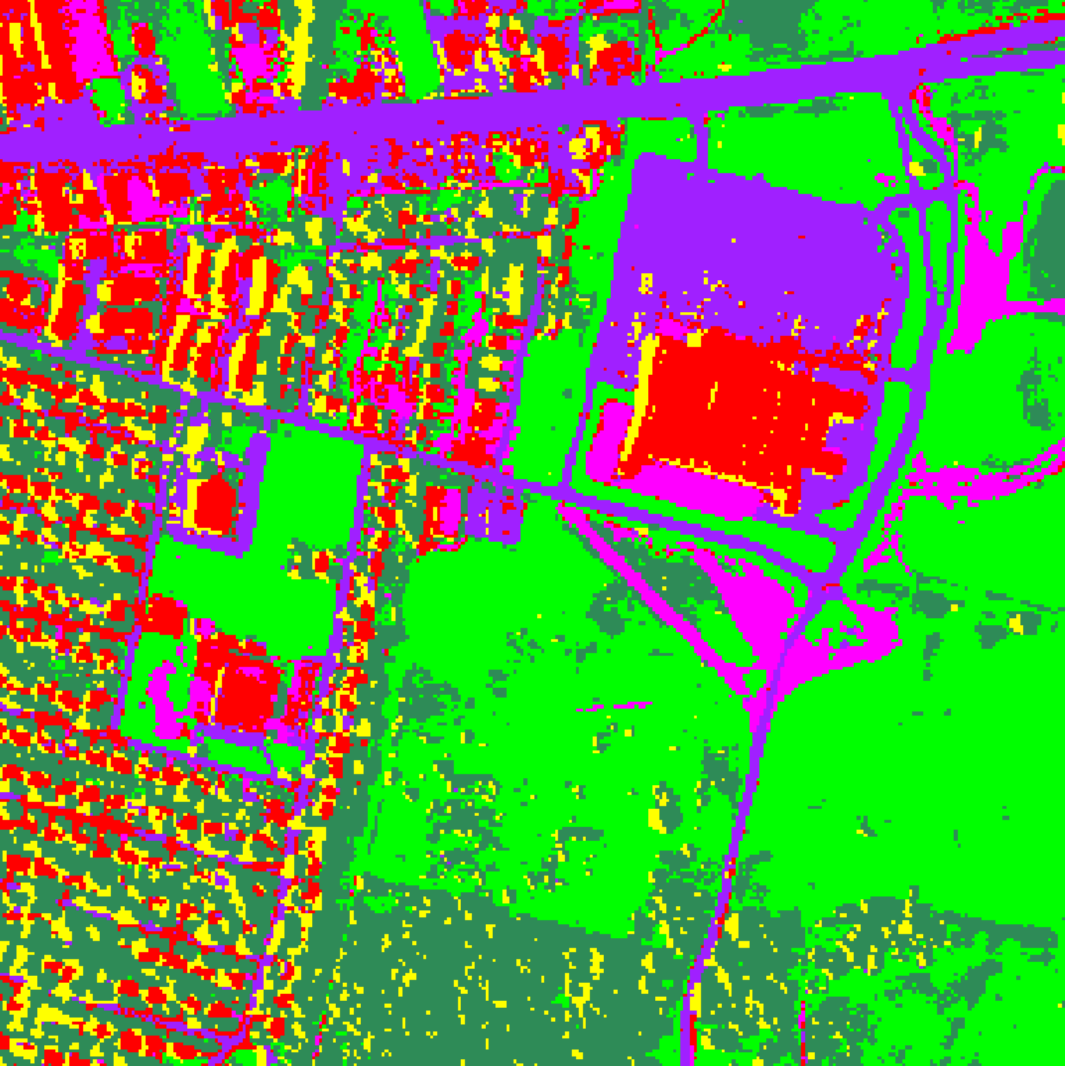, height = 1.2 in, keepaspectratio}}
\subfigure[]{\label{fig3f}
\epsfig{file = 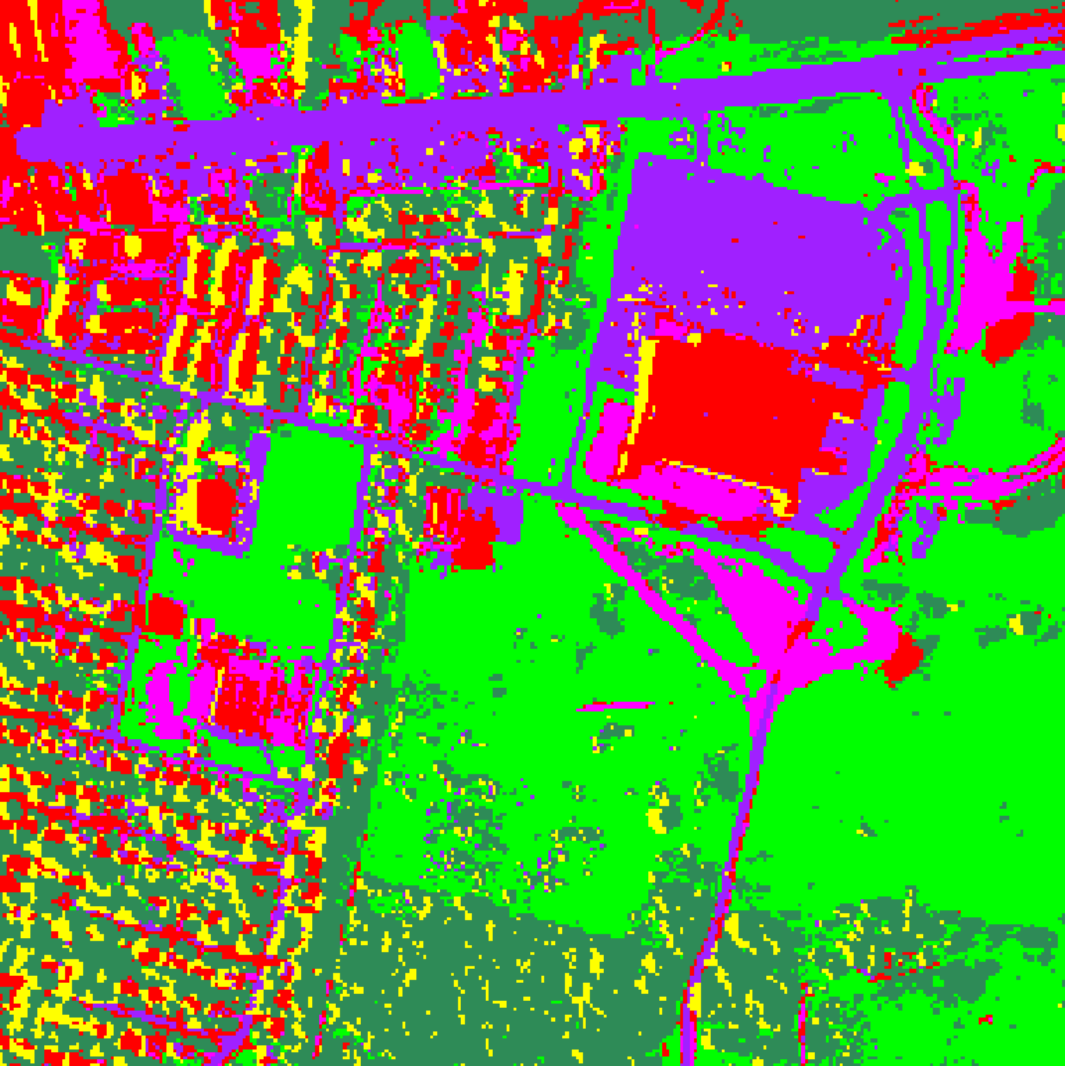, height = 1.2 in, keepaspectratio}}
\subfigure[]{\label{fig3g}
\epsfig{file = 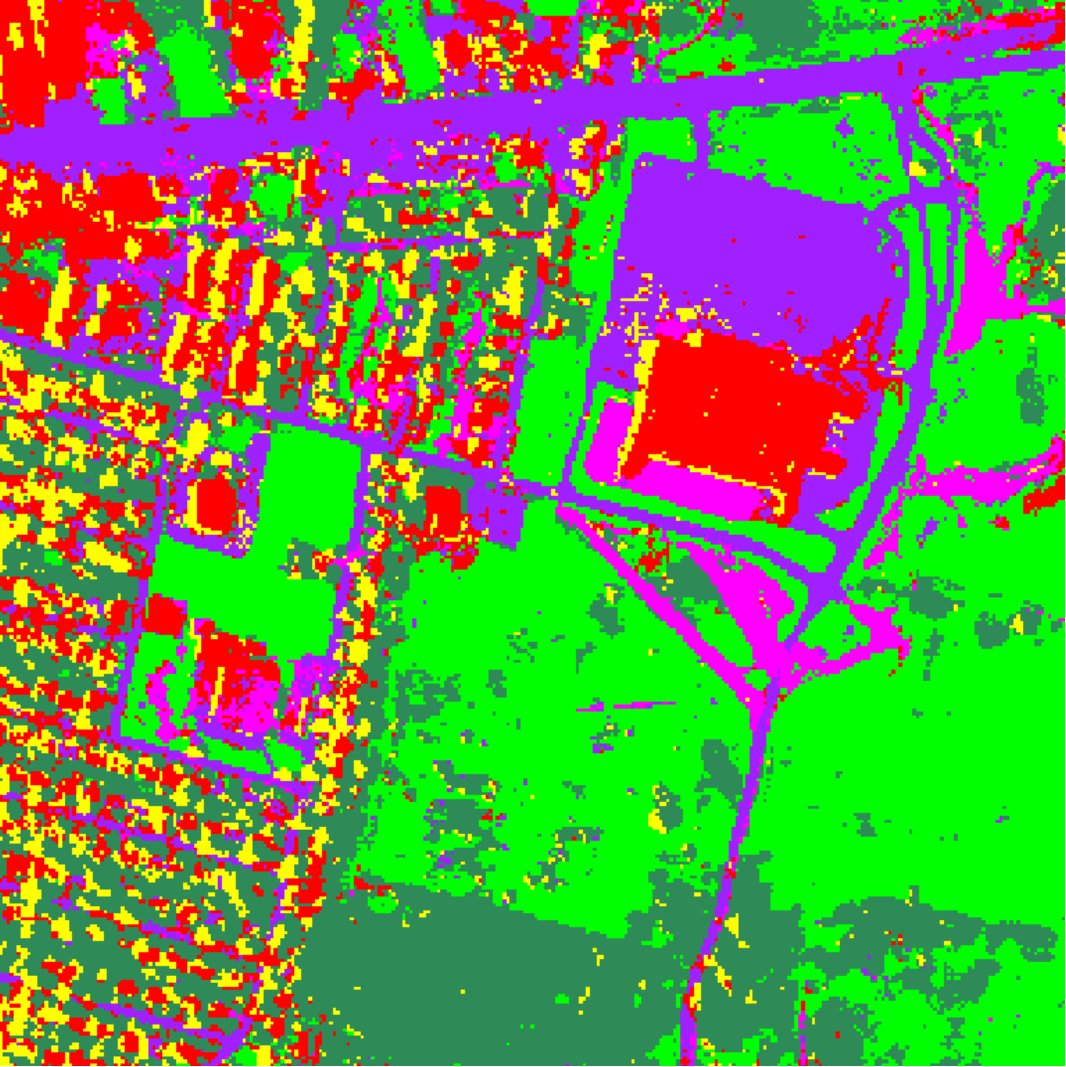, height = 1.2 in, keepaspectratio}}
\subfigure[]{\label{fig3h}
\epsfig{file = 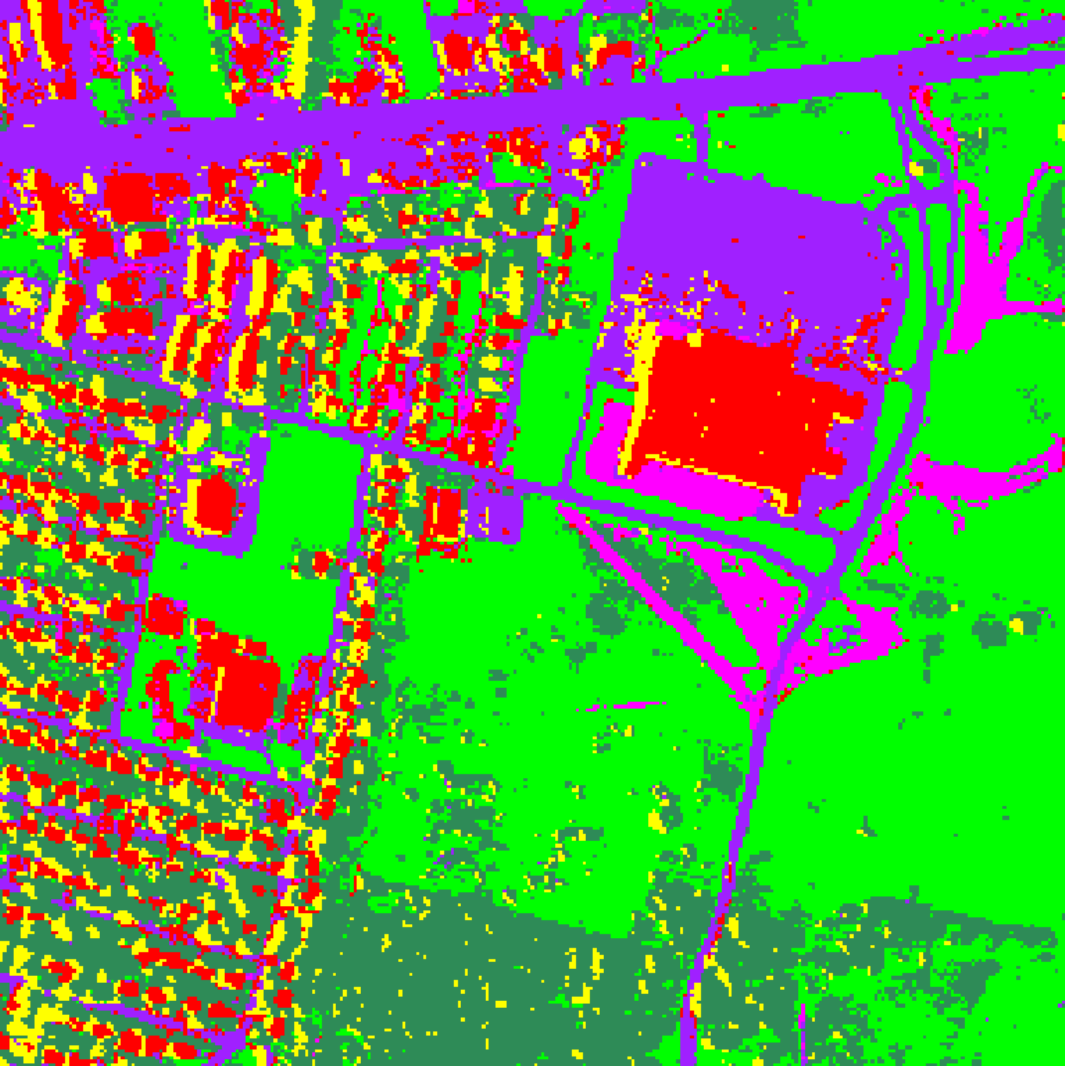, height = 1.2 in, keepaspectratio}}
\caption{ Classification maps with different feature representation methods of the HYDICE urban dataset. (a) Baseline, (b) SPCA, (c) SDA, (d) CNFE, (e) DNP, (f) CoLGP, (g) MFC, and (h) S3FSE.}
\label{fig3}
\end{center}
\end{figure*}

    \textbf{Classification results.} Figs. \ref{fig3a} to \ref{fig3h} show the classification maps of different feature representations based on SVM. In this experiment, the training samples in accordance with Fig. \ref{fig2c} are randomly chosen from the reference data and the rest of all are used as test samples. The feature dimensionalities of baseline and SDA are 327 and 5, respectively, while others are fixed at 50. At first glance, we learn that pixels of a few classes have mixed with each other in Figs. \ref{fig3a} to \ref{fig3g}. However, there are only a small number of mixed pixels existing in Fig. \ref{fig3h}. In summary, very little misclassifications could be observed in the classification map obtained by the S3FSE algorithm. These observations suggest that our proposed method could effectively and efficiently employ the spectral and spatial features for hyperspectral image classification.

\begin{figure}[htb]
 \centering
     \includegraphics[height = 1.7 in, keepaspectratio]{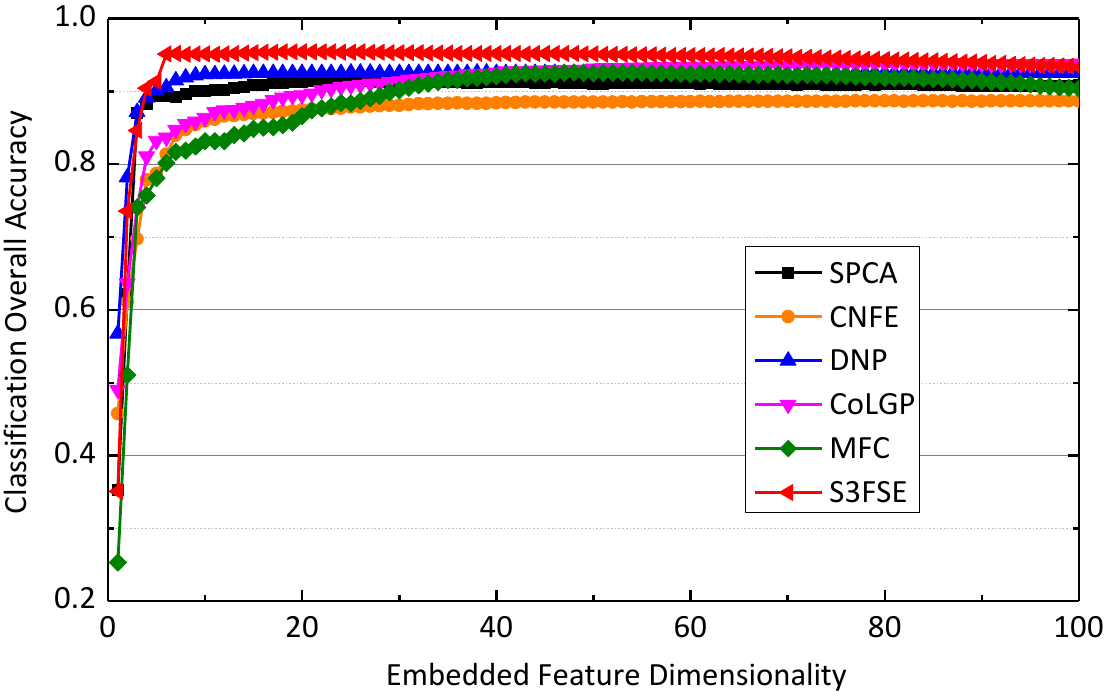}
      \caption{Embedded feature dimensionality \emph{d} respects to OA of the HYDICE urban dataset.}
      \label{fig4}
\end{figure}

    To further study the effectiveness of different feature representation methods, the detailed mean and standard deviation of classification rates in $10$ independent experiments are reported in Table \ref{tab1}.  In this table, some observations could be derived. To begin with, our proposed method achieves the best overall accuracy (OA) and Kappa coefficient. Furthermore, our proposed method obtains the highest classification accuracy in most of the individual classes. Finally, compare with CoLGP, the experimental results indicate that feature selection which discards noises and redundant features in S3FSE does great help to boost the discriminability of the learned feature representation. The computational costs of all the methods have also been reported in this table, we could learn that the proposed algorithm spends a little more time than its competitors, since an iteration step is required for the S3FSE optimization. For a more detailed comparison of the feature dimensionality reduction methods, the means of classification OA with regard to the reduced subspace dimensionalities from 1 to 100 have summarized in Fig. \ref{fig4}, this figure suggests that our proposed method achieves the best performance when the subspace dimensionality is larger than five. In addition, our proposed method remains a stable and leading performance when the dimensionality is increased.

\begin{table*}[tbp]\scriptsize
  \begin{center}
  \caption{Class-Specific Accuracies in Percentage of the HYDICE urban dataset}
  \label{tab1}
  \begin{tabular}{c c c c c c c c c}  \hline
  Classes            &Baseline          &SPCA                 &SDA                &CNFE              &DNP                    &CoLGP             &MFC                &S3FSE           \\ \hline
  Roof               &$92.02\pm6.68$    &$79.35\pm3.05$       &$85.08\pm3.91$     &$84.71\pm3.71$    &$72.71\pm6.05$         &$86.61\pm2.72$    &$88.40\pm2.68$     &$89.63\pm2.25$ \\
  Shadow             &$83.86\pm5.05$    &$89.15\pm1.74$       &$88.18\pm2.92$     &$83.67\pm3.90$    &$97.31\pm0.68$         &$90.09\pm2.99$    &$85.76\pm4.14$     &$91.82\pm1.56$\\
  Asphalt road       &$86.08\pm3.94$    &$93.04\pm2.27$       &$92.96\pm4.87$     &$91.62\pm2.40$    &$97.59\pm0.28$         &$94.78\pm1.08$    &$93.10\pm1.75$     &$96.72\pm0.45$\\
  Concrete road      &$91.73\pm6.96$    &$95.78\pm0.73$       &$95.34\pm1.36$     &$89.56\pm5.78$    &$96.26\pm0.64$         &$94.17\pm1.34$    &$93.95\pm2.07$     &$96.88\pm0.73$\\
  Grass              &$87.87\pm4.09$    &$95.71\pm2.01$       &$97.51\pm1.60$     &$91.10\pm4.27$    &$98.88\pm0.37$         &$97.48\pm1.20$    &$97.54\pm1.41$     &$98.22\pm2.37$\\
  Tree               &$77.96\pm5.13$    &$93.21\pm1.99$       &$94.27\pm2.50$     &$87.38\pm5.75$    &$91.80\pm2.18$         &$94.01\pm2.18$    &$94.15\pm1.66$     &$96.08\pm1.43$\\      \hline

  OA                 &$86.73\pm1.11$    &$91.07\pm0.84$       &$92.36\pm1.78$     &$88.55\pm2.95$    &$92.54\pm1.31$         &$93.15\pm0.67$    &$92.51\pm0.74$     &$95.13\pm0.81$\\
  Kappa              &$0.8385\pm0.01$   &$0.8909\pm0.01$      &$0.9067\pm0.02$    &$0.8603\pm0.03$   &$0.9088\pm0.01$        &$0.9163\pm0.01$   &$0.9086\pm0.01$    &$0.9405\pm0.01$\\    \hline
  Running time       &$3.1401s$     &$1.8862s$             &$0.5768s$           &$2.1781s$          &$1.9693s$               &$0.8027s$          &$1.1563s$           &$5.8434s$\\    \hline
\end{tabular}
\end{center}
\end{table*}

\begin{figure}[htbp]
\begin{center}
\subfigure[]{\label{fig5a}
\epsfig{file = 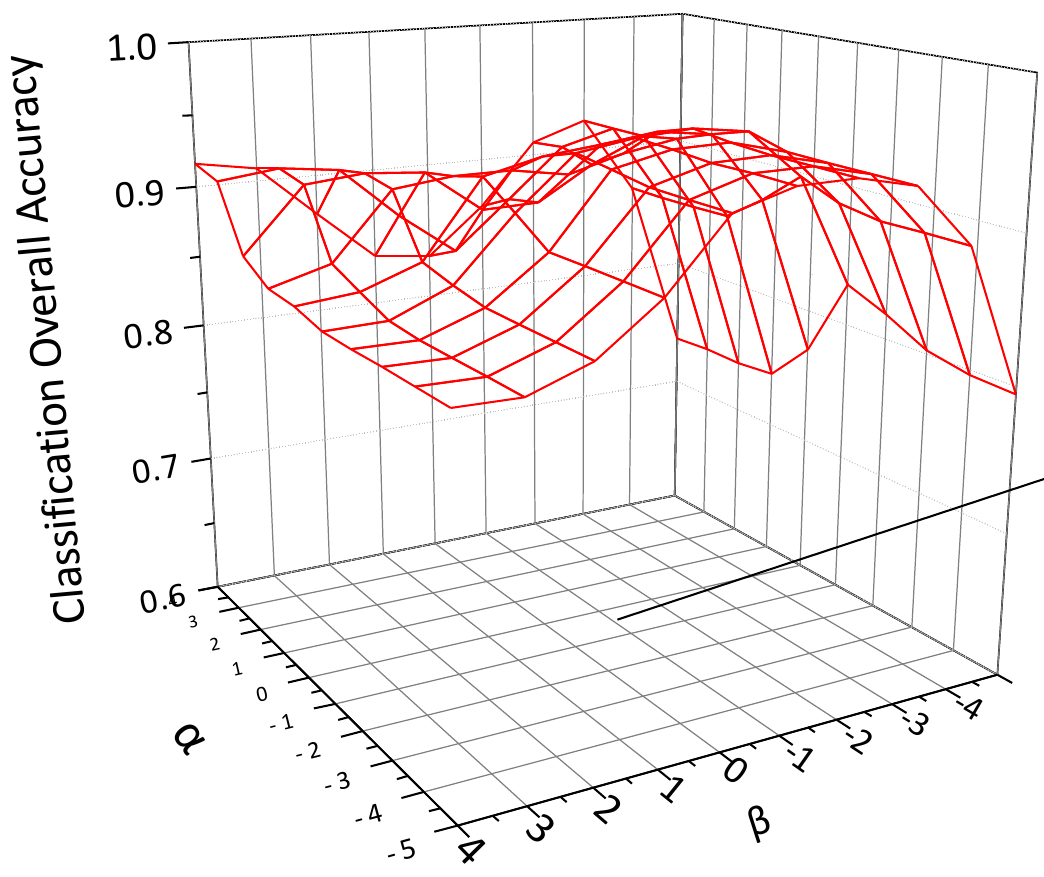, width = 1.6 in, keepaspectratio}}
\subfigure[]{\label{fig5b}
\epsfig{file = 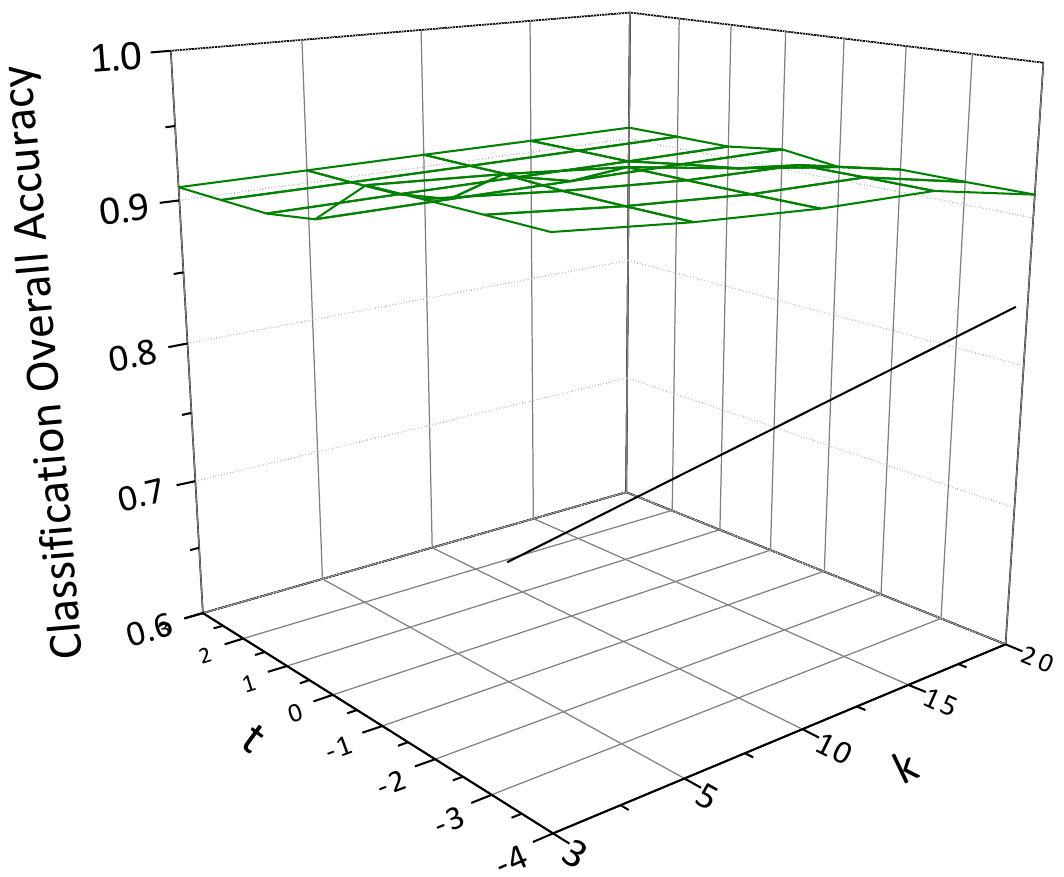, width = 1.6 in, keepaspectratio}}
\caption{Parameter sensitivity of the HYDICE urban dataset, (a) regularization parameters $\alpha$ and $\beta$ respect to OA, and (b) parameters \emph{k} and \emph{t} respect to OA.}
\label{fig5}
\end{center}
\end{figure}

    \textbf{Parameter sensitivity.} In the proposed S3FSE algorithm, there are several parameters, e.g., neighbor size $k$, kernel width $t$, regularization parameters $\alpha$ and $\beta$, to be decided in advance. Here we firstly study the effect of parameters $\alpha$ and $\beta$. In detail, we fix $k=5$ and $t=1$ as mentioned in the implementation details above, while the $\alpha$ and $\beta$ are set at various values of $\{10^{-5},10^{-4},\cdots,10^{4}\}$, respectively. Then classification OA respects to these two parameters is reported in Fig. \ref{fig5a}. It is obvious that we can obtain a good performance in a wide range of $\alpha$ if $\beta$ has been fixed. Meanwhile, the results have also suggested that small $\beta$ (e.g., $10^{-5}$ and $10^{-4}$) degrades the performance. Because small value $\beta$ keeps too many redundancy and noisy features which would have a invert effect on the performance. Based on this figure, we could fix the parameters $\alpha$ and $\beta$ at the middle interval of the tuned range to achieve a satisfactory performance in practice. We then show the effect of parameters $k$ and $t$ by fix $\alpha$ and $\beta$ as $10^{-1}$ and $10^{-2}$, respectively, the results of which are plotted in Fig. \ref{fig5b}. In this experiment, we tune $k$ and $t$ in the candidate sets of $\{3, 5, 10, 15, 20\}$ and $\{10^{-4},10^{-3},\cdots,10^{3}\}$, respectively. It is obvious that these two parameters have less effect to the classification performance compare to the two regularization parameters discussed above, since the OA varies much slighter in this figure. Therefore, it is reasonable for us to empirically set their values when running the proposed S3FSE algorithm in practice.

\begin{figure}[htbp]
\begin{center}
\subfigure[]{\label{fig6a}
\epsfig{file = 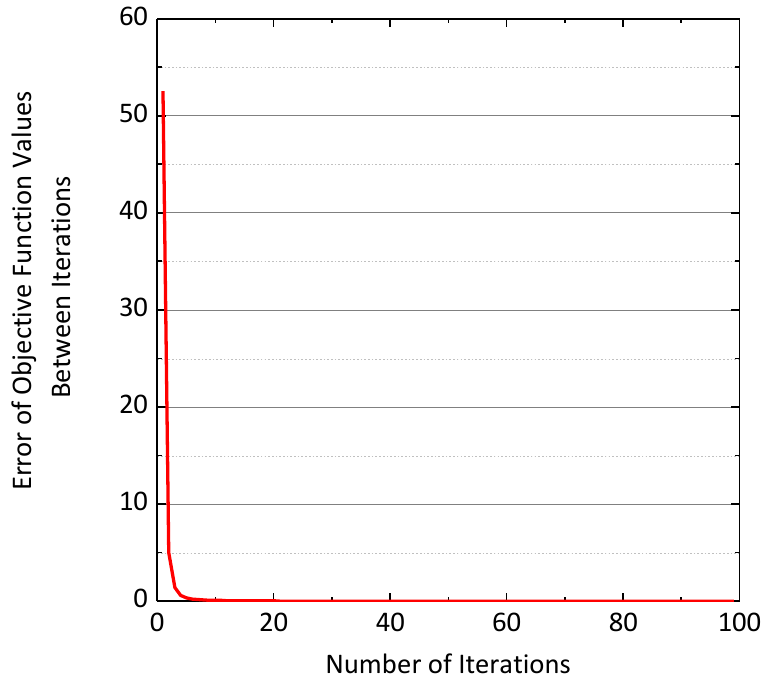, height = 1.5 in, keepaspectratio}}
\subfigure[]{\label{fig6b}
\epsfig{file = 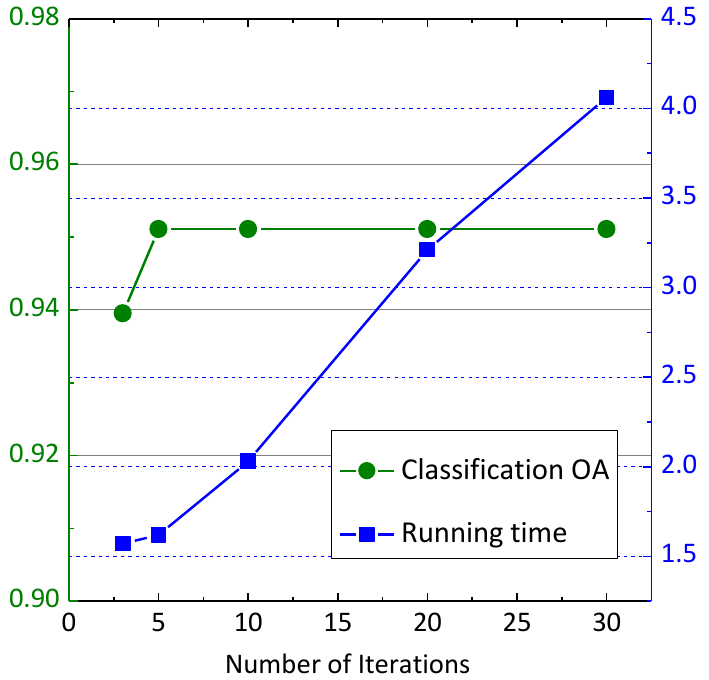, height = 1.5 in, keepaspectratio}}
\caption{Experimental convergence of the S3FSE algorithm on the HYDICE urban dataset, (a) the error of objective function values between iterations, and (b) classification overall accuracies and computational costs with different iterations.}
\label{fig6}
\end{center}
\end{figure}

    \textbf{Convergence study.} To study the convergence characteristics of the S3FSE optimization, we plot the error of objective function values (Eq. \eqref{formula25}) between iterations in Fig. \ref{fig6a}. As seen from this figure, it is clear that the proposed algorithm converges fast within less than ten iterations. We have also enforced the maxima number of iterations of the S3FSE optimization varies from 3 to 30 to see the classification performance, the results are presented in Fig. \ref{fig6b}. It is clear that the classification OA reaches to a stable value quickly when the number of iterations arrives at five, while the corresponding running time is keeping increases linearly. These observations demonstrate that our proposed optimization for S3FSE is highly efficient for the hyperspectral images classification task.

\subsection{Experiment 2:  HYDICE Washington DC dataset}

\begin{figure}[htbp]
\begin{center}
\subfigure[]{\label{fig7a}
\epsfig{file = 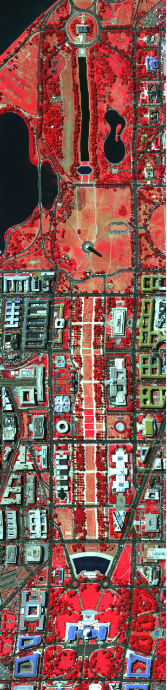, width = 0.7 in, keepaspectratio}}
\subfigure[]{\label{fig7b}
\epsfig{file = 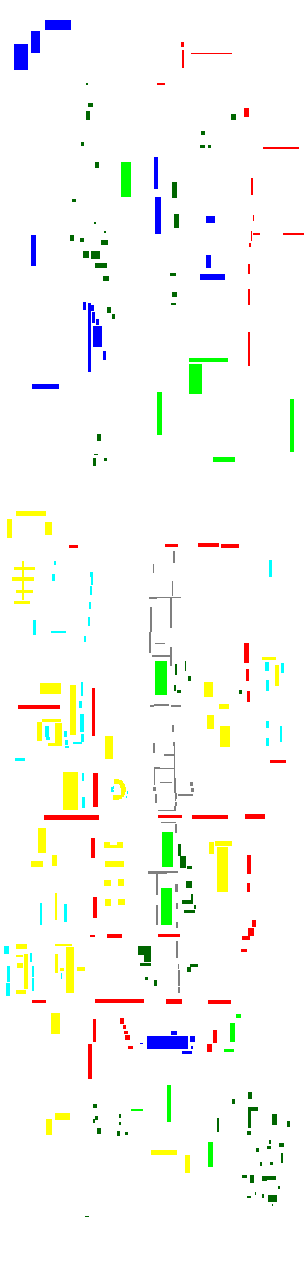, width = 0.7 in, keepaspectratio}}
\subfigure[]{\label{fig7c}
\epsfig{file = 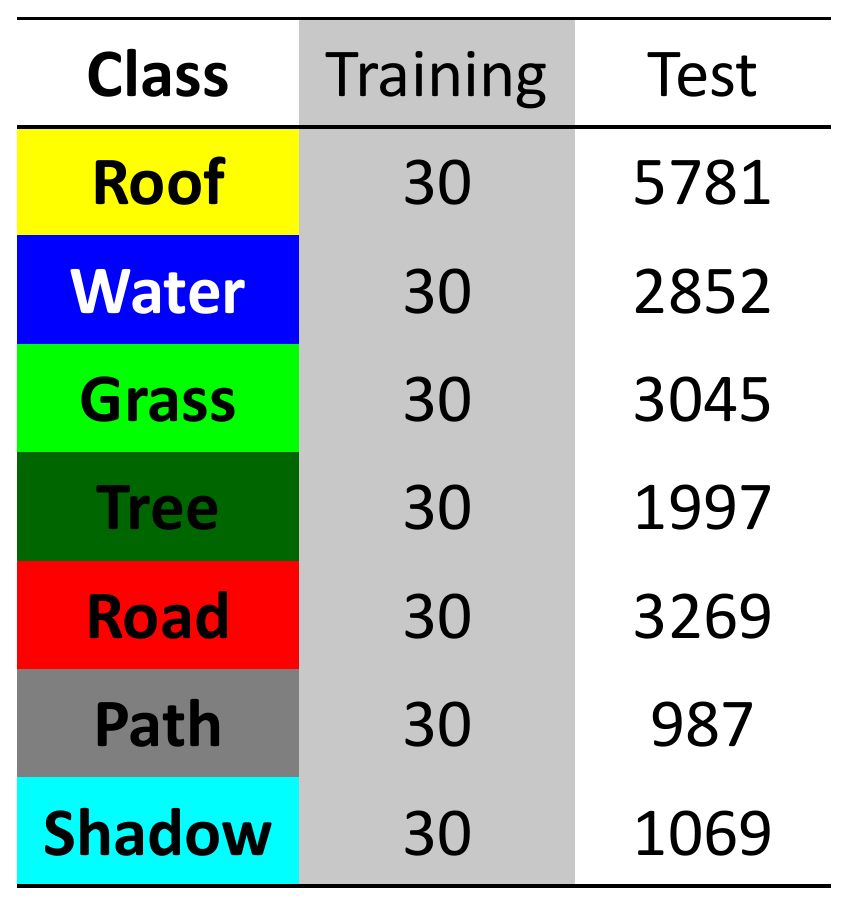, width = 1 in, keepaspectratio}}
\caption{ (a) The HYDICE Washington DC dataset, (b) the ground truth map, and (c) the number of training and test samples for classification.}
\label{fig7}
\end{center}
\end{figure}

    The HYDICE Washington DC dataset and its reference map are shown in Figs. \ref{fig7a} and \ref{fig7b}. This dataset is also an urban scene with high spectral and spatial resolutions. As given in Fig. \ref{fig7c}, there are seven significant land cover classes: roof, water, grass, tree, road, path, and shadow. It is also not an easy task to analysis this dataset, because some class pairs are very spectral similar, e.g., grass and tree, roof and road. As regard to the spectral and spatial features extraction, we have $191$-D spectral feature, $60$-D texture feature and $80$-D DMP feature for each pixel in this dataset. To investigate the performance of our proposed method under few training samples available condition, we have also randomly selected $30$ samples of each class as training samples and considered the rest of reference data as test samples. The detailed number of training samples and test samples is given in Fig. \ref{fig7c}.

\begin{figure*}[htbp]
\begin{center}
\subfigure[]{\label{fig8a}
\epsfig{file = 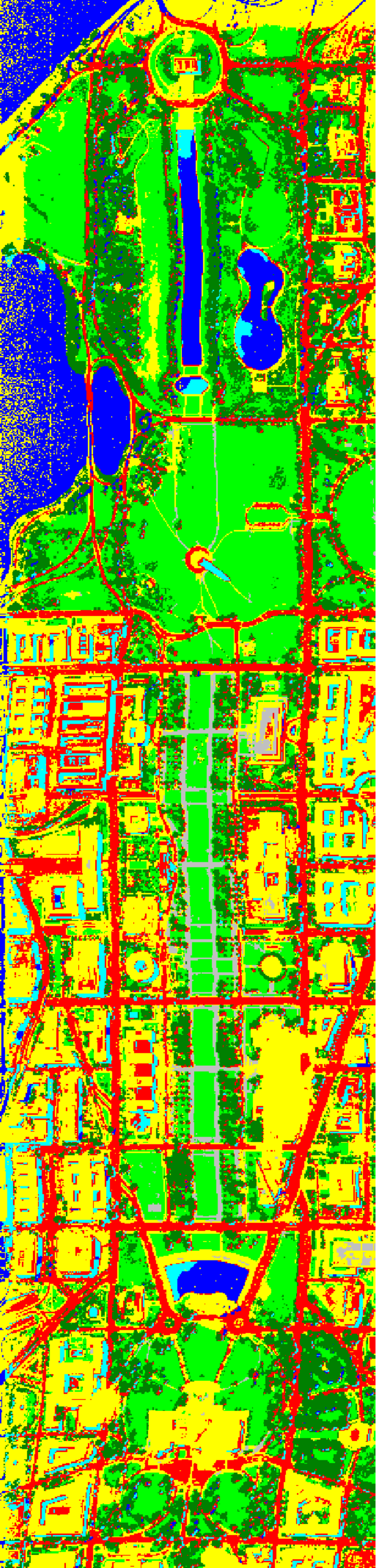, width = 0.7 in, keepaspectratio}}
\subfigure[]{\label{fig8b}
\epsfig{file = 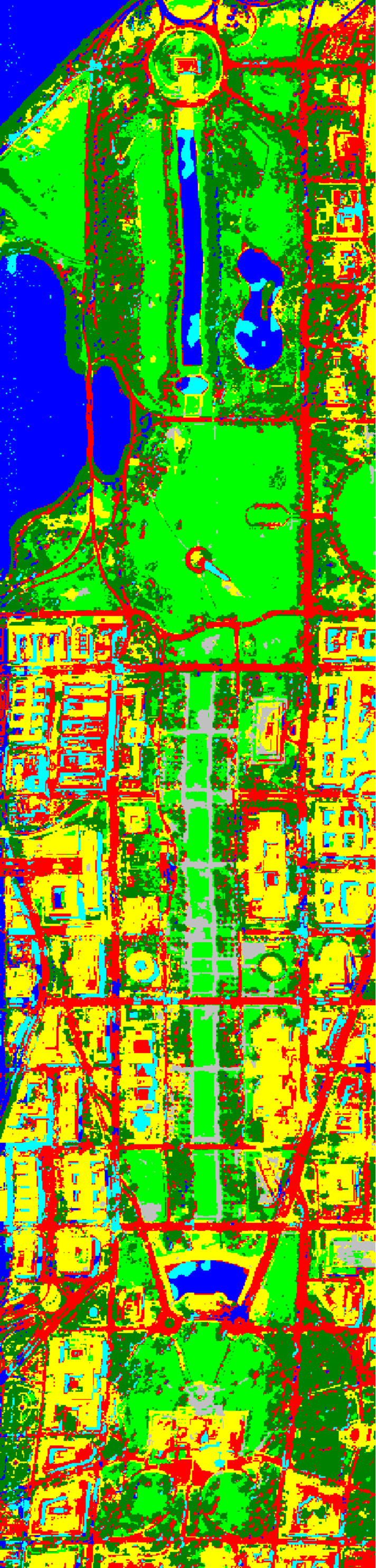, width = 0.7 in, keepaspectratio}}
\subfigure[]{\label{fig8c}
\epsfig{file = 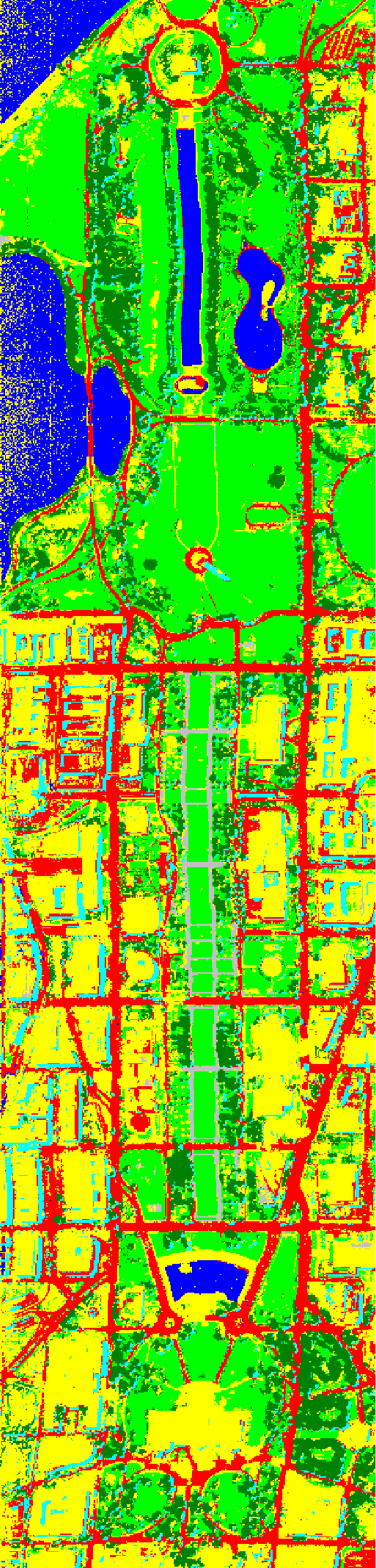, width = 0.7 in, keepaspectratio}}
\subfigure[]{\label{fig8d}
\epsfig{file = 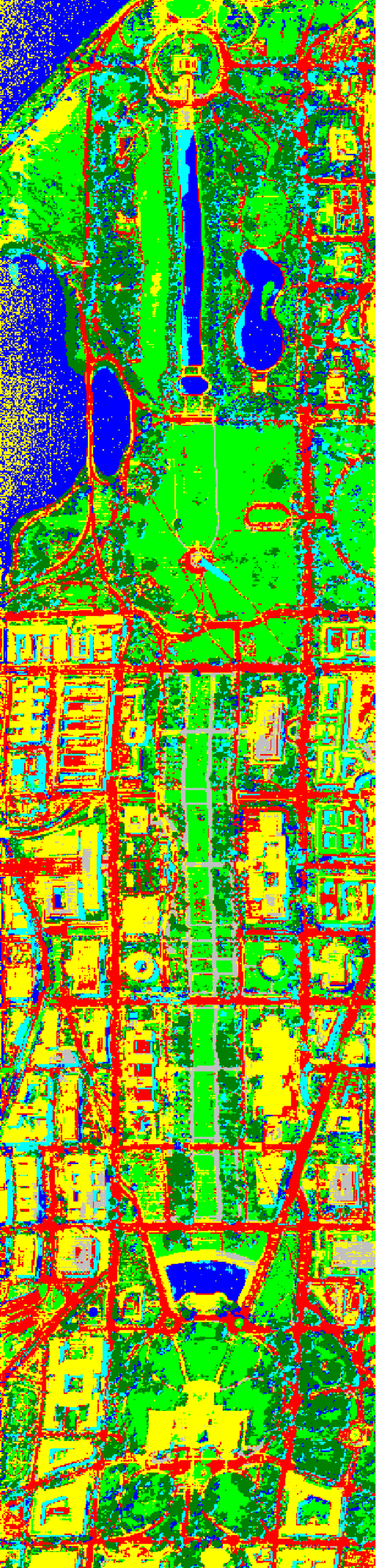, width = 0.7 in, keepaspectratio}}
\subfigure[]{\label{fig8e}
\epsfig{file = 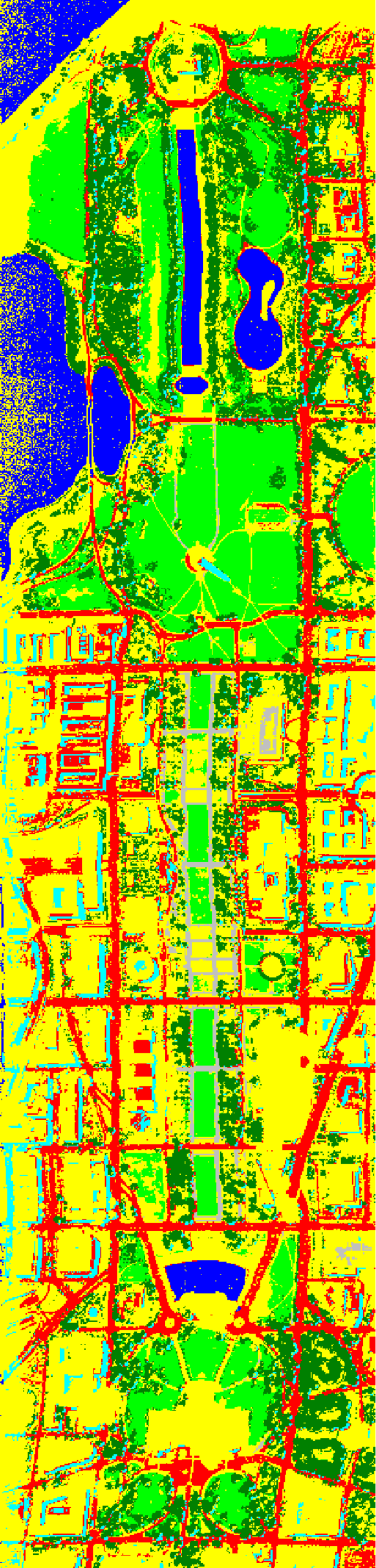, width = 0.7 in, keepaspectratio}}
\subfigure[]{\label{fig8f}
\epsfig{file = 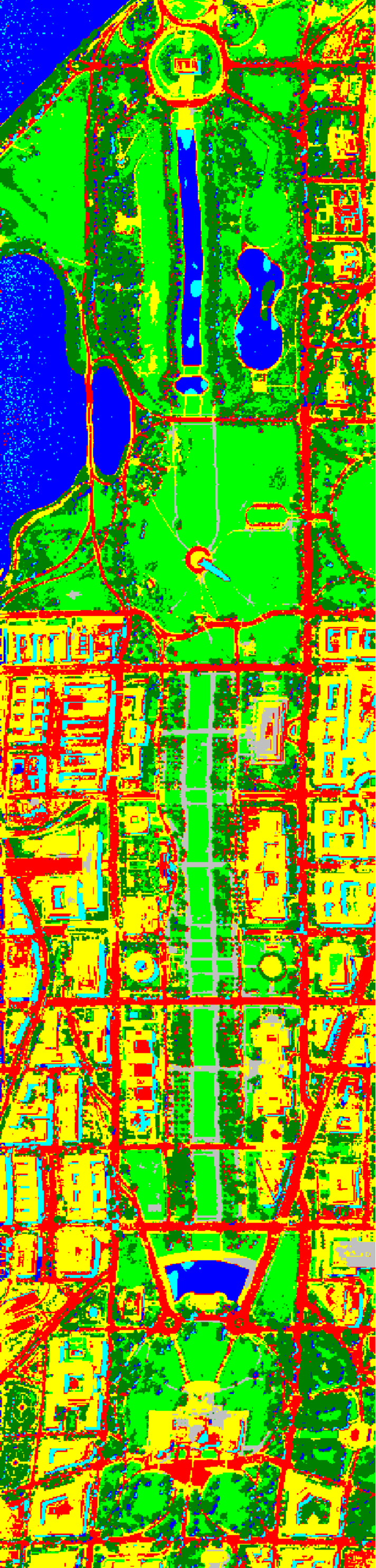, width = 0.7 in, keepaspectratio}}
\subfigure[]{\label{fig8g}
\epsfig{file = 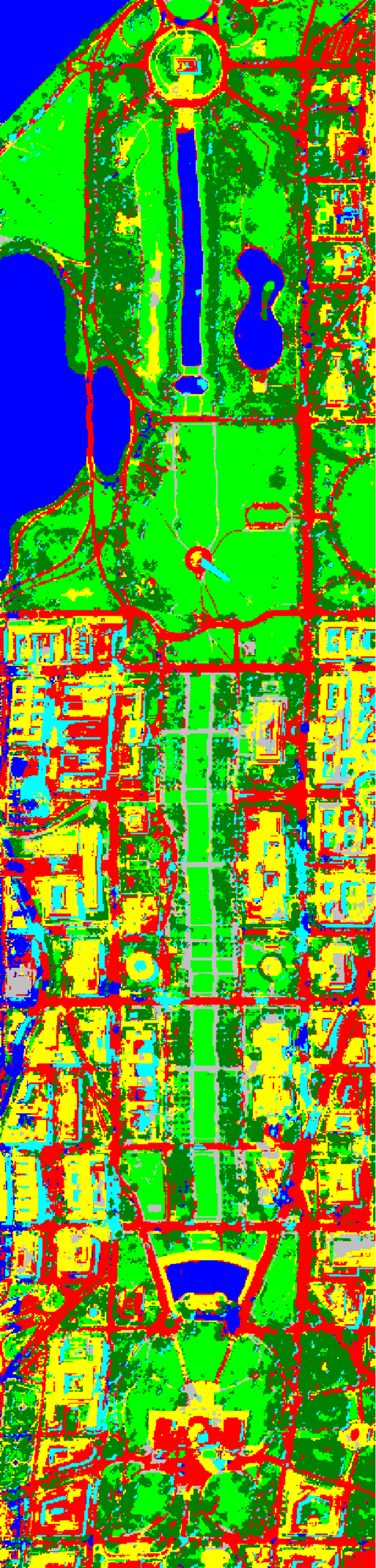, width = 0.7 in, keepaspectratio}}
\subfigure[]{\label{fig8h}
\epsfig{file = 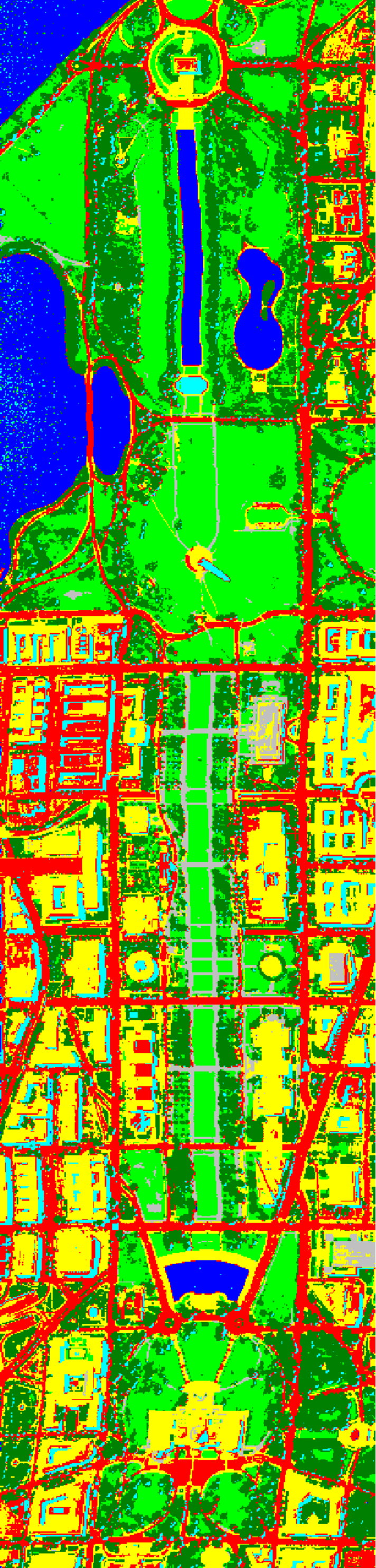, width = 0.7 in, keepaspectratio}}
\caption{ Classification maps with different feature representation methods of the Washington DC dataset. (a) Baseline, (b) SPCA, (c) SDA, (d) CNFE, (e) DNP, (f) CoLGP, (g) MFC, and (h) S3FSE.}
\label{fig8}
\end{center}
\end{figure*}

\begin{figure}[htb]
 \centering
     \includegraphics[height = 1.7 in, keepaspectratio]{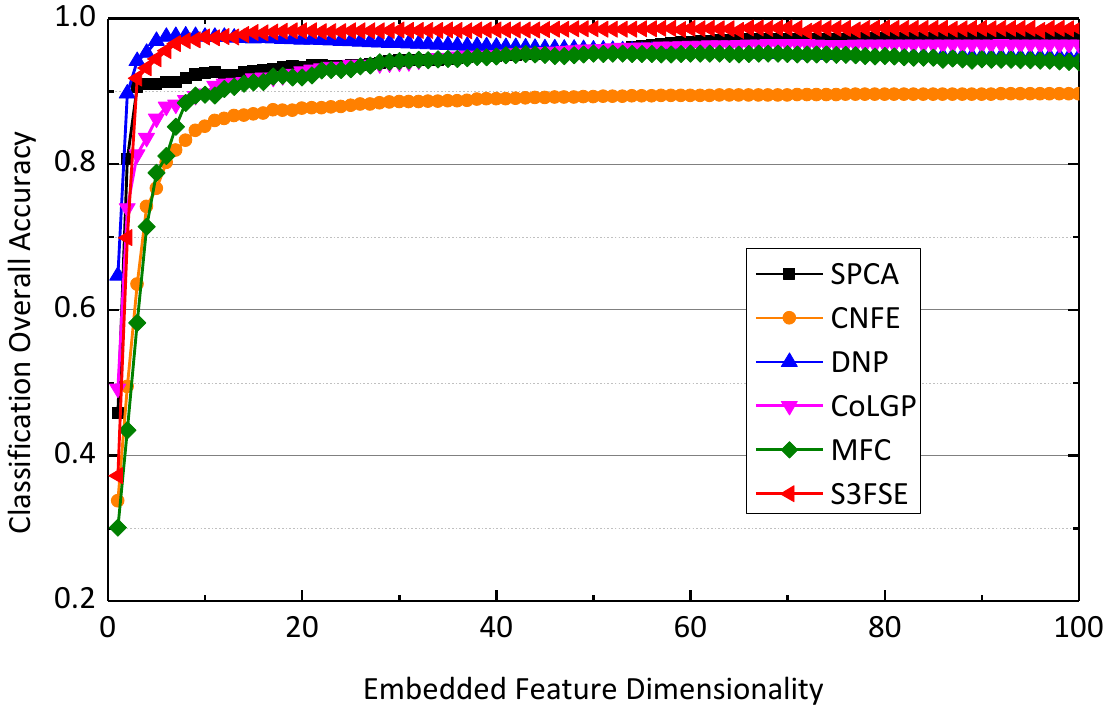}
      \caption{Embedded feature dimensionality \emph{d} respects to OA of the Washington DC dataset.}
      \label{fig9}
 \end{figure}

    The eight feature representation based classification maps of the HYDICE Washington DC dataset are presented in Figs. \ref{fig8a} to \ref{fig8h}. Among them, the dimensionalities of baseline and SDA are $331$ and $6$, respectively, while the number of low dimensional feature subspaces of other approaches is unified to $50$. As shown in Fig. \ref{fig8}, the proposed method achieves the best result in visual interpretation, just as the experimental results discussed in the HYDICE urban dataset above. For example, it is clear that roof pixels exist in road pixels in all classification maps. Nevertheless, very little roof pixels have been mixed into the road pixels in Fig. \ref{fig8h}, which shows the result of our proposed approach. For a more detailed accuracy assessment of all the methods, we have also reported the mean and standard deviation of 10 group of independent classification overall accuracies in Table \ref{tab2}, as well as the curves of OA with regard to various $d$ from 1 to 100 in Fig. \ref{fig9}, all of these statistics indicate that our proposed S3FSE algorithm has achieved the best classification performance.

\begin{table*}[tbp] \scriptsize
  \begin{center}
  \setlength{\arrayrulewidth}{0.5pt}
  \caption{Class-Specific Accuracies in Percentage of the HYDICE Washington DC Dataset}
  \label{tab2}
  \begin{tabular}{c c c c c c c c c} \hline
  Classes   &Baseline          &SPCA                   &SDA               &CNFE               &DNP                 &CoLGP                 &MFC                  &S3FSE  \\  \hline
  Roof      &$97.37\pm1.36$    &$95.10\pm1.65$         &$95.30\pm5.52$    &$85.80\pm5.81$     &$99.59\pm0.35$      &$93.65\pm1.84$        &$93.24\pm2.90$       &$97.56\pm0.67$\\
  Water     &$91.15\pm2.89$    &$95.48\pm2.91$         &$96.36\pm1.61$    &$87.71\pm5.24$     &$93.04\pm1.53$      &$95.83\pm3.61$        &$98.62\pm1.20$       &$98.88\pm0.90$\\
  Grass     &$98.73\pm1.28$    &$98.86\pm1.57$         &$97.57\pm2.62$    &$93.13\pm5.15$     &$96.47\pm3.81$      &$99.42\pm0.46$        &$97.51\pm3.86$       &$99.98\pm0.30$\\
  Tree      &$97.87\pm0.57$    &$97.66\pm0.92$         &$97.84\pm0.73$    &$95.48\pm2.01$     &$96.51\pm1.35$      &$98.92\pm0.44$        &$98.40\pm1.25$       &$99.71\pm0.20$\\
  Road      &$89.92\pm3.19$    &$90.43\pm3.71$         &$95.62\pm1.90$    &$88.07\pm3.36$     &$90.42\pm2.10$      &$93.67\pm3.02$        &$90.43\pm4.54$       &$97.81\pm1.18$\\
  Path      &$96.21\pm1.61$    &$98.49\pm0.74$         &$97.01\pm1.28$    &$91.75\pm4.41$     &$94.42\pm2.95$      &$97.60\pm1.18$        &$97.02\pm1.98$       &$99.08\pm0.45$\\
  Shadow    &$95.32\pm3.02$    &$96.27\pm3.29$         &$94.83\pm2.50$    &$91.05\pm3.09$     &$96.70\pm1.81$      &$95.87\pm2.50$        &$97.03\pm1.29$       &$98.39\pm0.38$\\  \hline
  OA        &$95.25\pm0.86$    &$95.47\pm0.65$         &$96.21\pm1.55$    &$89.27\pm3.08$     &$95.77\pm0.89$      &$95.79\pm1.03$        &$95.20\pm1.29$       &$98.54\pm0.29$\\
  Kappa     &$0.9413\pm0.01$   &$0.9443\pm0.01$        &$0.9534\pm0.02$   &$0.8690\pm0.03$    &$0.9476\pm0.01$     &$0.9484\pm0.01$       &$0.9411\pm0.02$      &$0.9821\pm0.01$\\    \hline
  Running time       &$4.7564s$     &$2.1638s$             &$0.9484s$           &$2.8409s$          &$2.6278s$               &$1.0219s$          &$1.7394s$           &$6.7367s$\\    \hline
\end{tabular}
\end{center}
\end{table*}

\subsection{Experiment 3: Pavia city ROSIS dataset}

\begin{figure*}[htbp]
\begin{center}
\subfigure[]{\label{fig11a}
\epsfig{file = 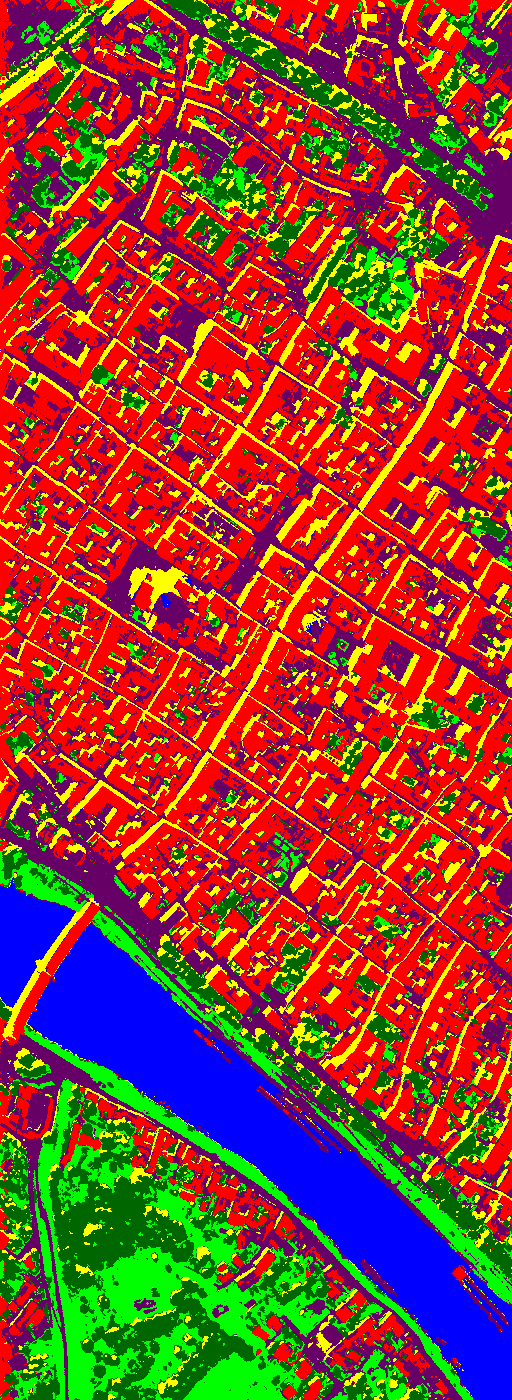, width = 0.8 in, keepaspectratio}}
\subfigure[]{\label{fig11b}
\epsfig{file = 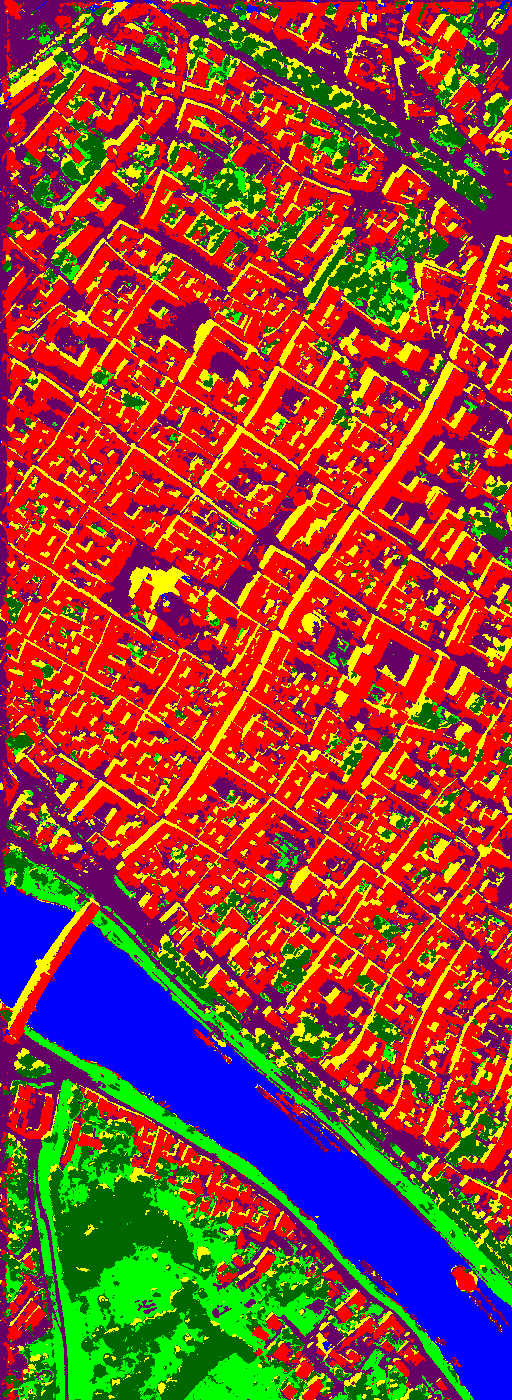, width = 0.8 in, keepaspectratio}}
\subfigure[]{\label{fig11c}
\epsfig{file = 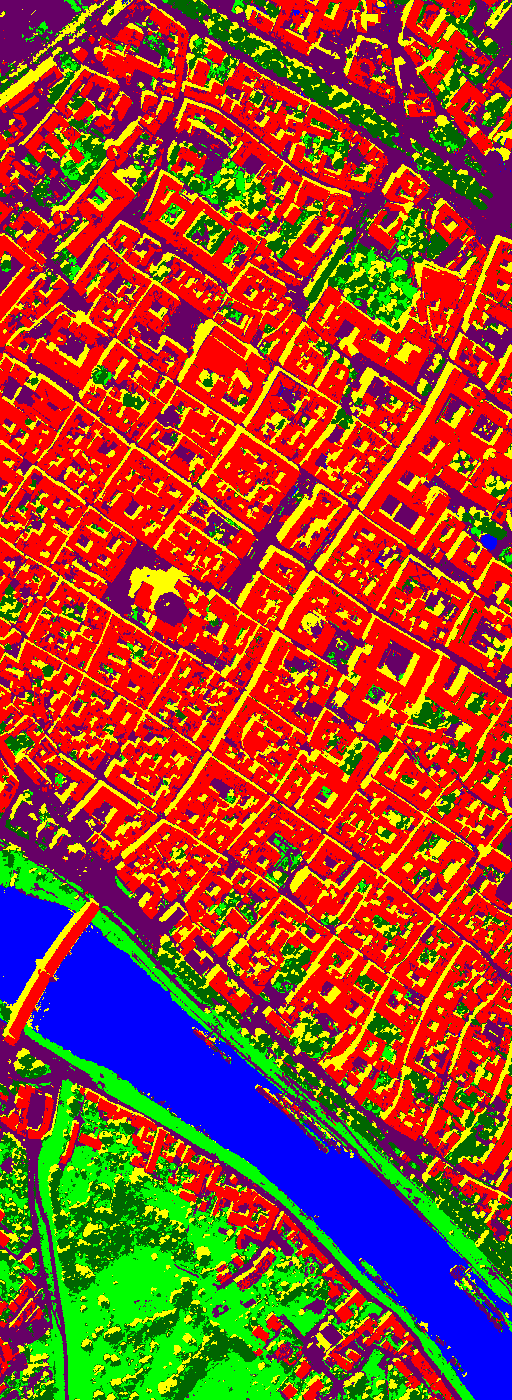, width = 0.8 in, keepaspectratio}}
\subfigure[]{\label{fig11d}
\epsfig{file = 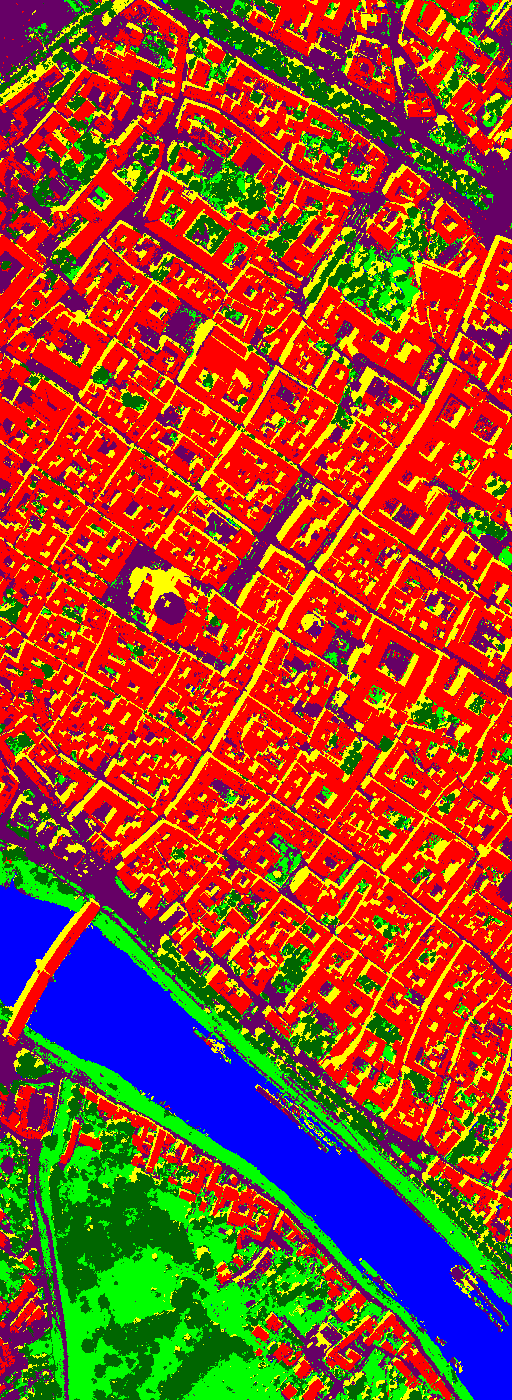, width = 0.8 in, keepaspectratio}}
\subfigure[]{\label{fig11e}
\epsfig{file = 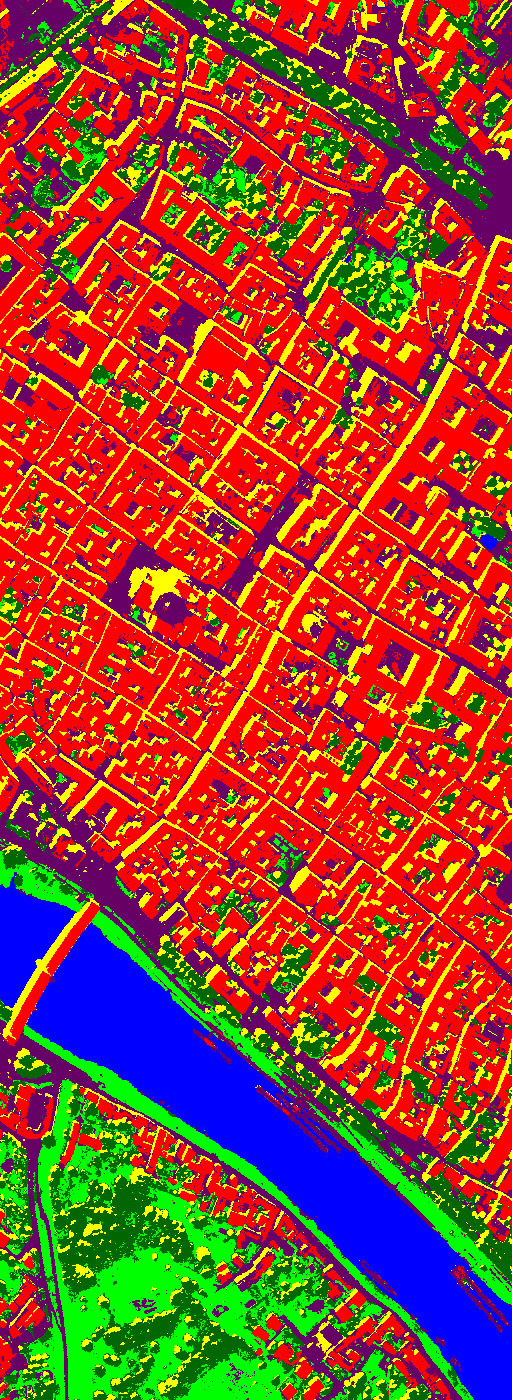, width = 0.8 in, keepaspectratio}}
\subfigure[]{\label{fig11f}
\epsfig{file = 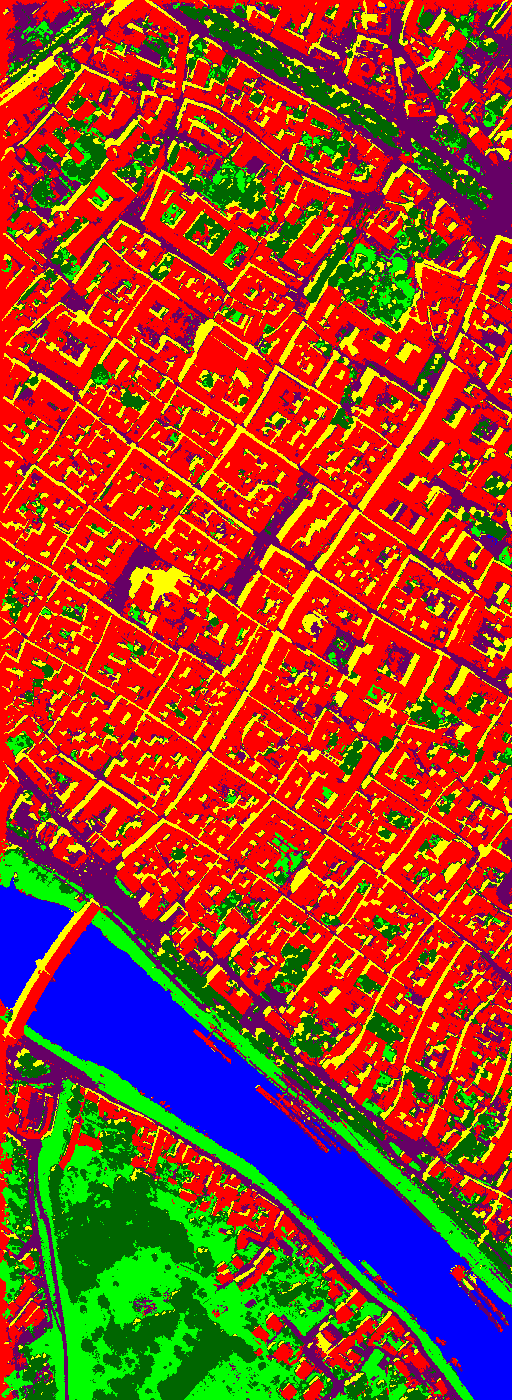, width = 0.8 in, keepaspectratio}}
\subfigure[]{\label{fig11g}
\epsfig{file = 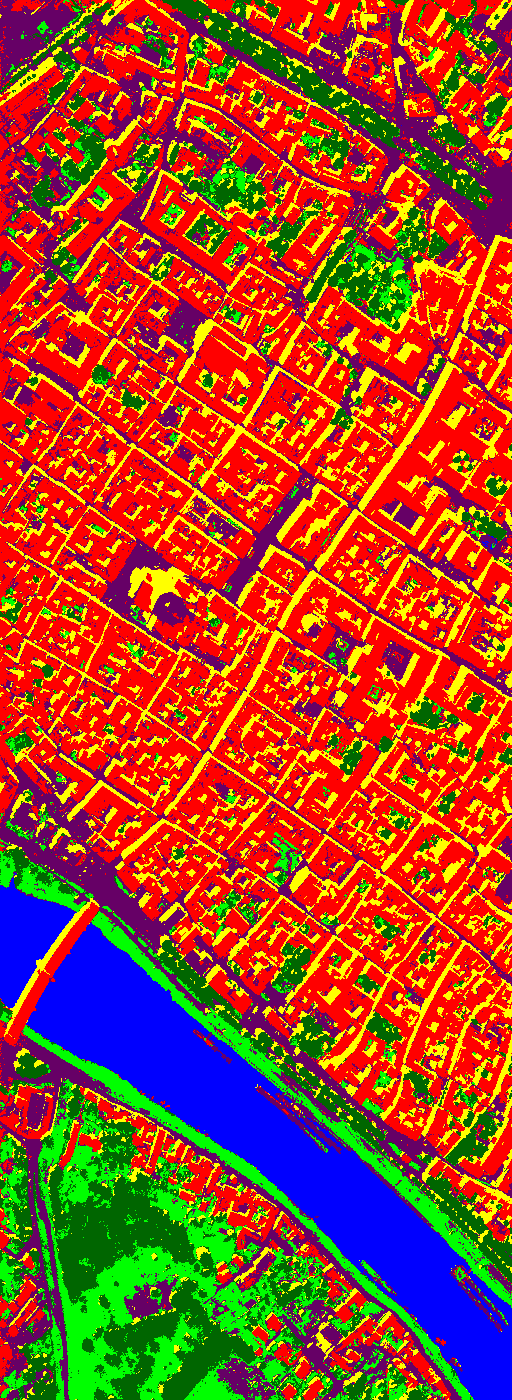, width = 0.8 in, keepaspectratio}}
\subfigure[]{\label{fig11h}
\epsfig{file = 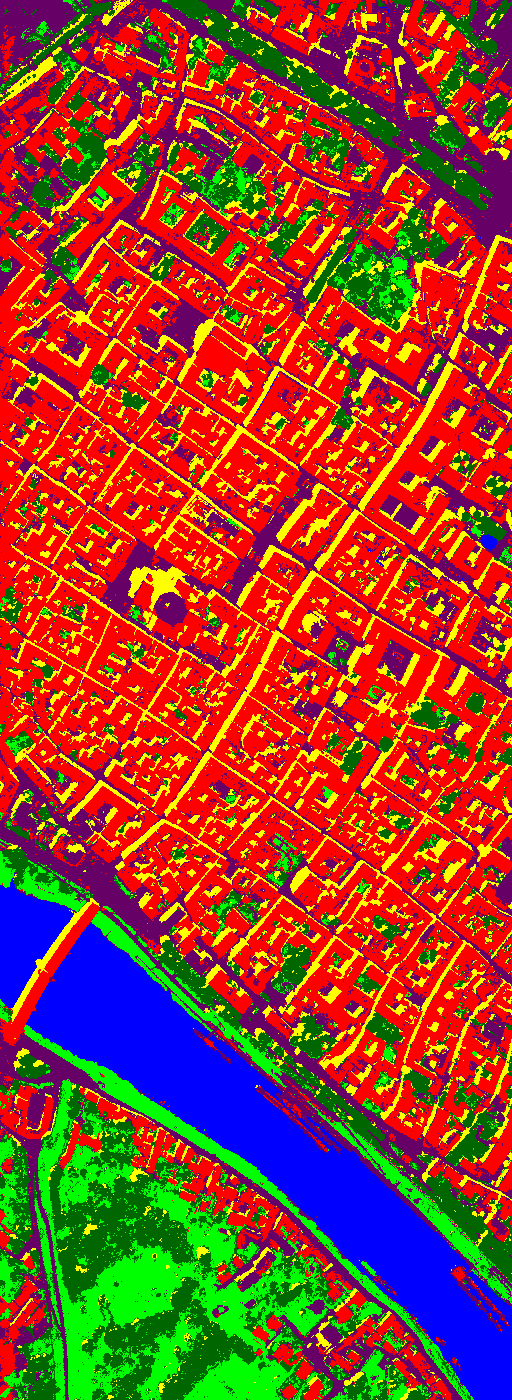, width = 0.8 in, keepaspectratio}}
\caption{ Classification maps with different feature representation methods of the ROSIS Pavia city dataset. (a) Baseline, (b) SPCA, (c) SDA, (d) CNFE, (e) DNP, (f) CoLGP, (g) MFC, and (h) S3FSE.}
\label{fig11}
\end{center}
\end{figure*}

\begin{figure}[htbp]
\begin{center}
\subfigure[]{\label{fig10a}
\epsfig{file = 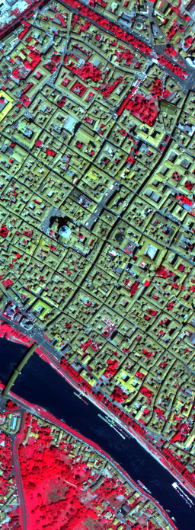, width = 0.8 in, keepaspectratio}}
\subfigure[]{\label{fig10b}
\epsfig{file = 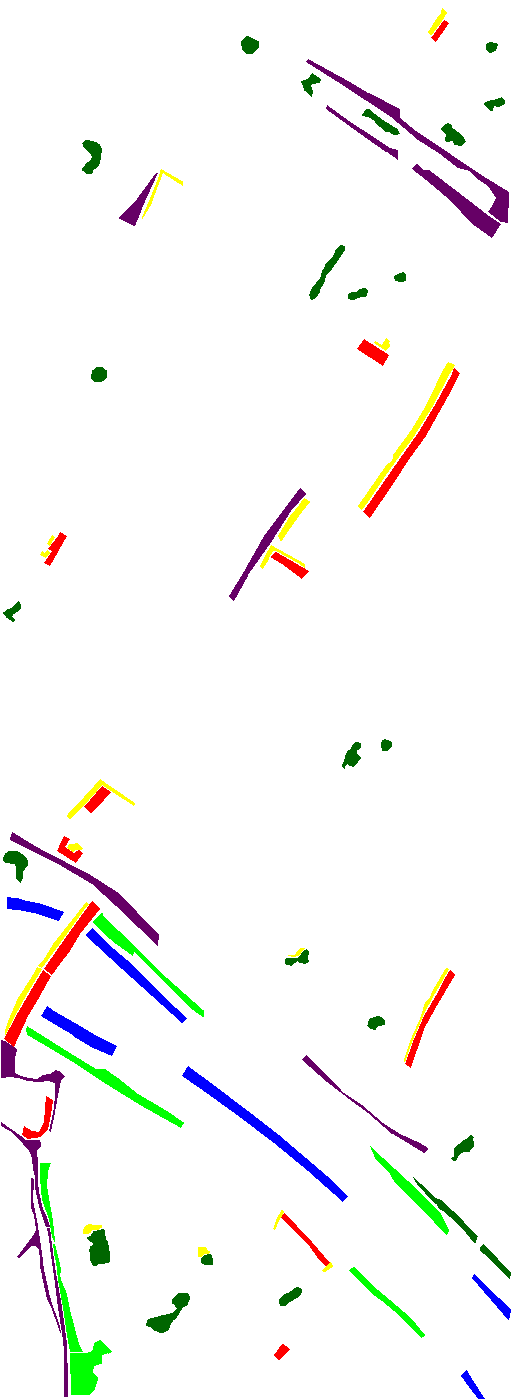, width = 0.8 in, keepaspectratio}}
\subfigure[]{\label{fig10c}
\epsfig{file = 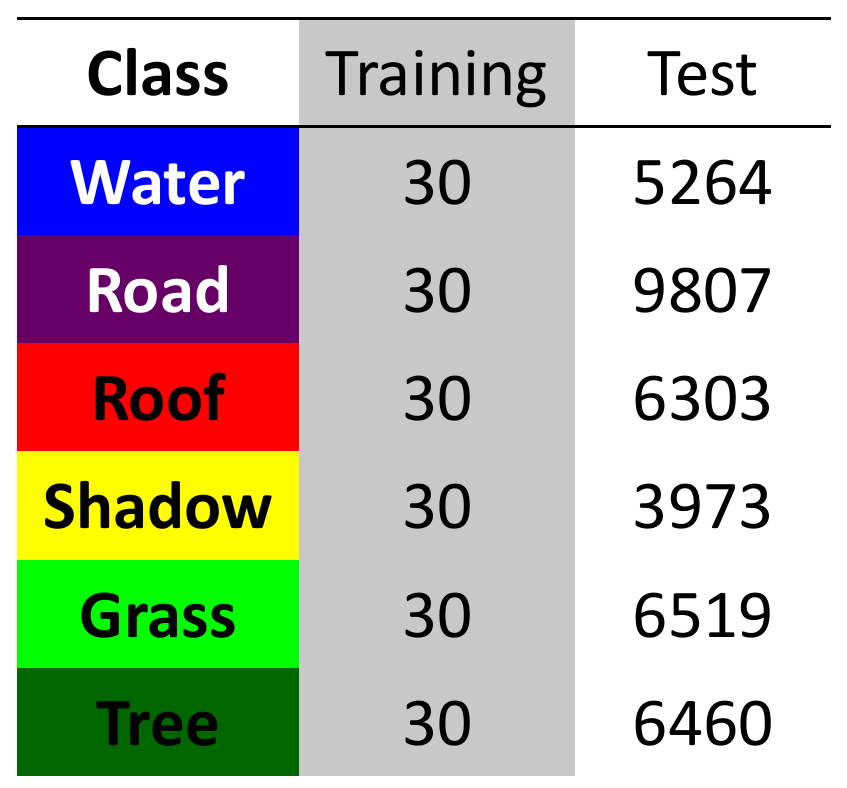, width = 1 in, keepaspectratio}}
\caption{ (a) The ROSIS Pavia city dataset, (b) the ground truth map, and (c) the number of training and test samples for classification.}
\label{fig10}
\end{center}
\end{figure}

\begin{figure}[htb]
 \centering
     \includegraphics[height = 1.7 in, keepaspectratio]{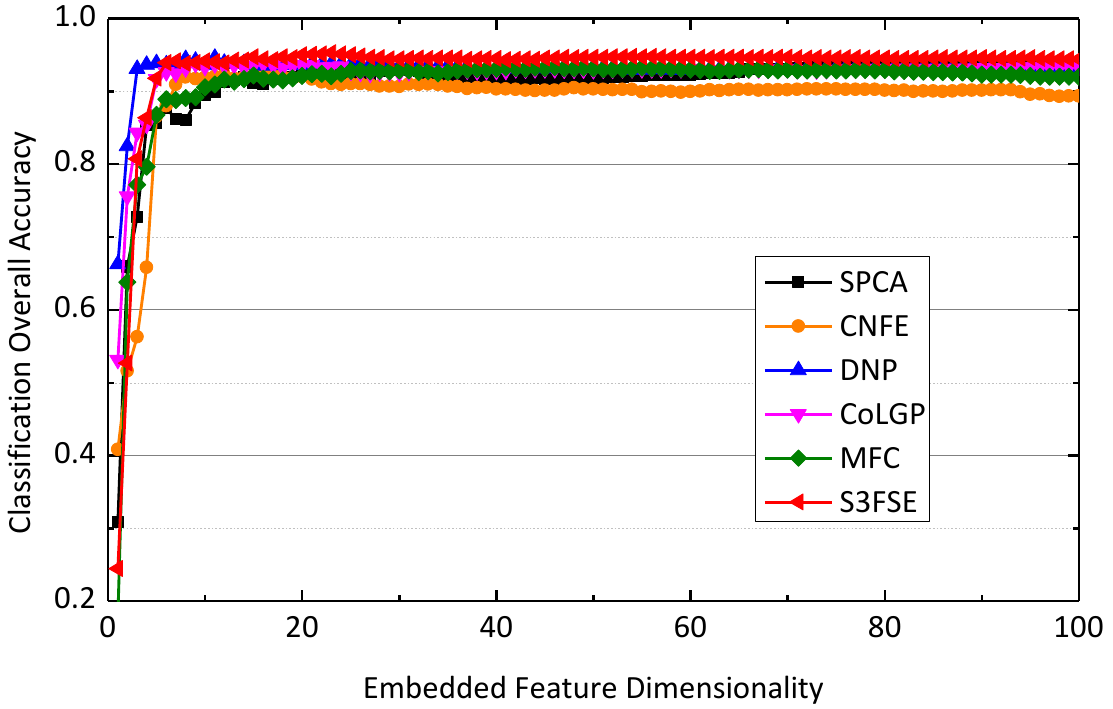}
      \caption{Embedded feature dimensionality \emph{d} respects to OA of the ROSIS Pavia city dataset.}
      \label{fig12}
 \end{figure}

  The Pavia city ROSIS dataset and the reference map are shown in Figs. \ref{fig10a} and \ref{fig10b}. According to the reference data, there are six classes of samples that need to be analyzed, i.e., water, road, roof, shadow, grass, and tree. In this dataset, we still randomly selected $30$ samples per class as training sample and the reminding for testing, the numbers of training and test samples are listed in Fig. \ref{fig10c}. The classification maps, the detailed classification accuracy statistics, and the performance of OA with regard to various $d$ have shown in Fig. \ref{fig11}, Table \ref{tab3}, and Fig. \ref{fig12}, respectively. In this experiment, the detailed parameter setting of all the algorithms is in accordance with the instructions described in the experiment setup subsection. Similarly to the results reported in the previous datasets, we observe that the proposed S3FSE algorithm achieves the best classification performance from both visual interpretation and accuracy. Consequently, the consistent conclusions reported in the aforementioned three hyperspectral datasets demonstrate that the proposed S3FSE algorithm is an effective approach for hyperspectral images spectral and spatial features representation and classification.

\begin{table*}[tbp] \scriptsize
  \begin{center}
  \setlength{\arrayrulewidth}{0.5pt}
  \caption{Class-Specific Accuracies in Percentage of the ROSIS Pavia city dataset}
  \label{tab3}
  \begin{tabular}{c c c c c c c c c} \hline
  Classes   &Baseline          &SPCA                   &SDA               &CNFE               &DNP                 &CoLGP                 &MFC                  &S3FSE  \\  \hline
  Water      &$94.34\pm1.48$    &$99.73\pm1.09$         &$99.69\pm0.73$    &$98.80\pm2.43$     &$99.85\pm0.32$      &$99.47\pm0.52$        &$99.90\pm0.58$       &$99.89\pm0.18$\\
  Road       &$90.16\pm1.78$    &$92.00\pm2.55$         &$95.15\pm1.90$    &$88.53\pm4.12$     &$95.08\pm1.90$      &$92.76\pm2.87$        &$94.38\pm2.11$       &$96.41\pm1.37$\\
  Roof       &$87.44\pm1.87$    &$90.42\pm1.77$         &$86.90\pm0.79$    &$84.68\pm2.42$     &$88.60\pm1.75$      &$91.77\pm2.18$        &$88.47\pm0.76$       &$90.18\pm2.09$\\
  Shadow     &$95.92\pm1.96$    &$88.44\pm2.14$         &$90.74\pm2.97$    &$91.02\pm3.58$     &$95.94\pm2.34$      &$91.61\pm2.21$        &$90.54\pm3.41$       &$92.42\pm2.68$\\
  Grass      &$90.40\pm3.23$    &$91.54\pm2.13$         &$94.71\pm4.25$    &$93.60\pm3.79$     &$91.66\pm2.71$      &$95.98\pm2.48$        &$96.84\pm2.50$       &$98.55\pm2.65$\\
  Tree       &$90.34\pm2.51$    &$88.49\pm2.07$         &$77.45\pm4.42$    &$88.13\pm2.52$     &$90.54\pm1.84$      &$86.94\pm1.89$        &$88.30\pm1.52$       &$89.66\pm2.23$\\   \hline
  OA         &$90.51\pm0.89$    &$91.76\pm0.80$         &$90.90\pm1.36$    &$90.36\pm3.22$     &$93.41\pm0.74$      &$92.96\pm1.03$        &$93.17\pm0.87$       &$94.68\pm0.56$\\
  Kappa      &$0.8900\pm0.01$   &$0.8996\pm0.01$        &$0.8891\pm0.02$   &$0.8827\pm0.04$    &$0.9197\pm0.01$     &$0.9143\pm0.01$       &$0.9167\pm0.01$      &$0.9351\pm0.01$\\    \hline
  Running time       &$6.9188s$     &$2.0322s$             &$1.3628s$           &$2.9648s$          &$2.5773s$               &$1.6211s$          &$1.7949s$           &$4.6402s$\\    \hline
\end{tabular}
\end{center}
\end{table*}

\section{Conclusions}
    In this paper, we propose a simultaneous spectral-spatial feature selection and extraction (S3FSE) algorithm for hyperspectral images spectral-spatial feature representation and classification. In S3FSE, firstly, the spectral and spatial features of each pixel are extracted. Then, the geometric structure of the spectral and spatial features are preserved and the consistency and complementary information are exploited via the co-manifold learning and co-graph regularization. Finally, considering that some measurements are redundant or noisy, we impose $\ell_{2,1}$ norm to co-regularize the learned projection matrices to simultaneously performance the feature selection and feature extraction. As a result, the redundant features and noises have been  discarded, and only the significant original features have been transformed. The learned common low dimensional feature representation can be interpreted as a linear combination of a subset of significant original features. Extensive experiments on three public available hyperspectral datasets have been conducted to confirm the validity of the S3FSE algorithm. The experimental results show that S3FSE consistently outperforms the other comparison methods for hyperspectral images classification. For the future work, we will consider to extend the current S3FSE to its kernel version to better fit for the complex datasets in practice.

\section*{Acknowledgment}

    The authors would like to thank the handing editor and the anonymous reviewers for their careful reading and helpful remarks, which have contributed in improving the quality of this paper.

\ifCLASSOPTIONcaptionsoff
  \newpage
\fi

\bibliographystyle{IEEEtran}
\footnotesize
\bibliography{ref}

\begin{thebibliography}{10}
\providecommand{\url}[1]{#1}
\csname url@samestyle\endcsname
\providecommand{\newblock}{\relax}
\providecommand{\bibinfo}[2]{#2}
\providecommand{\BIBentrySTDinterwordspacing}{\spaceskip=0pt\relax}
\providecommand{\BIBentryALTinterwordstretchfactor}{4}
\providecommand{\BIBentryALTinterwordspacing}{\spaceskip=\fontdimen2\font plus
\BIBentryALTinterwordstretchfactor\fontdimen3\font minus
  \fontdimen4\font\relax}
\providecommand{\BIBforeignlanguage}[2]{{%
\expandafter\ifx\csname l@#1\endcsname\relax
\typeout{** WARNING: IEEEtran.bst: No hyphenation pattern has been}%
\typeout{** loaded for the language `#1'. Using the pattern for}%
\typeout{** the default language instead.}%
\else
\language=\csname l@#1\endcsname
\fi
#2}}
\providecommand{\BIBdecl}{\relax}
\BIBdecl

\bibitem{GD2014}
G.~Camps-Valls, D.~Tuia, L.~Bruzzone, and J.~A. Benediktsson, ``Advances in
  hyperspectral image classification: Earth monitoring with statistical
  learning methods,'' \emph{IEEE Signal Process. Mag.}, vol.~31, no.~1, pp.
  45--54, 2014.

\bibitem{LH2012}
L.~Zhang, H.~Shen, W.~Gong, and H.~Zhang, ``Adjustable model-based fusion
  method for multispectral and panchromatic images,'' \emph{IEEE Trans. Syst.
  Man Cybern. Part B Cybern.}, vol.~42, no.~6, pp. 1693--1704, 2012.

\bibitem{YL2012}
Y.~Zhong and L.~Zhang, ``An adaptive artificial immune network for supervised
  classification of multi-/hyperspectral remote sensing imagery,'' \emph{IEEE
  Trans. Geosci. Remote Sens.}, vol.~50, no.~3, pp. 894--909, 2012.

\bibitem{HX2015}
H.~Shen, X.~Li, Q.~Cheng, C.~Zeng, G.~Yang, H.~Li, and L.~Zhang, ``Missing
  information reconstruction of remote sensing data: A technical review,''
  \emph{IEEE Geosci. Remote Sens. Mag.}, vol.~3, no.~3, pp. 61--85, 2015.

\bibitem{MY2013}
M.~Fauvel, Y.~Tarabalka, J.~A. Benediktsson, J.~Chanussot, and J.~C. Tilton,
  ``Advances in spectral-spatial classification of hyperspectral images,''
  \emph{Proc. IEEE}, vol. 101, no.~3, pp. 566--569, 2013.

\bibitem{YY2016}
Y.~Zhou and Y.~Wei, ``Learning hierarchical spectral-spatial features for
  hyperspectral image classification,'' \emph{IEEE Trans. Cybern.}, vol.~46,
  no.~7, pp. 1667--1678, 2016.

\bibitem{WS2012}
W.~Li, S.~Prasad, J.~E. Fowler, and L.~M. Bruce, ``Locality-preserving
  dimensionality reduction and classification for hyperspectral image
  analysis,'' \emph{IEEE Trans. Geosci. Remote Sens.}, vol.~50, no.~4, pp.
  1185--1198, 2012.

\bibitem{HY2015}
H.~Yuan and Y.~Y. Tang, ``Learning with hypergraph for hyperspectral image
  feature extraction,'' \emph{IEEE Geosci. Remote Sens. Lett.}, vol.~12, no.~8,
  pp. 1695--1699, 2015.

\bibitem{YB2014}
Y.~Zhong, B.~Zhao, and L.~Zhang, ``Multiagent object-based classifier for high
  spatial resolution imagery,'' \emph{IEEE Trans. Geosci. Remote Sens.},
  vol.~52, no.~2, pp. 841--857, 2014.

\bibitem{JY2015}
J.~Zhao, Y.~Zhong, and L.~Zhang, ``Detail-preserving smoothing classifier based
  on conditional random fields for high spatial resolution remote sensing
  imagery,'' \emph{IEEE Trans. Geosci. Remote Sens.}, vol.~53, no.~5, pp.
  2440--2452, 2015.

\bibitem{LS2015}
L.~Fang, S.~Li, W.~Duan, J.~Ren, and J.~A. Benediktsson, ``Classification of
  hyperspectral images by exploiting spectral-spatial information of superpixel
  via multiple kernels,'' \emph{IEEE Trans. Geosci. Remote Sens.}, vol.~53,
  no.~12, pp. 6663--6674, 2015.

\bibitem{HY2016}
H.~Yuan and Y.~Y. Tang, ``Spectral-spatial shared linear regression for
  hyperspectral image classification,'' \emph{IEEE Trans. Cybern.}, DOI:
  10.1109/TCYB.2016.2533430, 2016.

\bibitem{JH2014}
J.~Li, H.~Zhang, Y.~Huang, and L.~Zhang, ``Hyperspectral image classification
  by nonlocal joint collaborative representation with a locally adaptive
  dictionary,'' \emph{IEEE Trans. Geosci. Remote Sens.}, vol.~52, no.~6, pp.
  3707--3719, 2014.

\bibitem{LS2014}
L.~Fang, S.~Li, X.~Kang, and J.~A. Benediktsson, ``Spectral-spatial
  hyperspectral image classification via multiscale adaptive sparse
  representation,'' \emph{IEEE Trans. Geosci. Remote Sens.}, vol.~52, no.~12,
  pp. 7738--7749, 2014.

\bibitem{HG2014}
H.~Li, G.~Xiao, T.~Xia, Y.~Y. Tang, and L.~Li, ``Hyperspectral image
  classification using functional data analysis,'' \emph{IEEE Trans. Cybern.},
  vol.~44, no.~9, pp. 1544--1555, 2014.

\bibitem{JY2016}
J.~Zhao, Y.~Zhong, H.~Shu, and L.~Zhang, ``High-resolution image classification
  integrating spectral-spatial-location cues by conditional random fields,''
  \emph{IEEE Trans. Image Process.}, vol.~25, no.~9, pp. 4033--4045, 2016.

\bibitem{XX2016}
X.~Guo, X.~Huang, L.~Zhang, L.~Zhang, A.~Plaza, and J.~A. Benediktsson,
  ``Support tensor machines for classification of hyperspectral remote sensing
  imagery,'' \emph{IEEE Trans. Geosci. Remote Sens.}, vol.~54, no.~6, pp.
  3248--3264, 2016.

\bibitem{YT2010}
Y.~Tarabalka, J.~Chanussot, and J.~A. Benediktsson, ``Segmentation and
  classification of hyperspectral images using minimum spanning forest grown
  from automatically selected markers,'' \emph{IEEE Trans. Syst. Man Cybern.
  Part B Cybern.}, vol.~40, no.~5, pp. 1267--1279, 2010.

\bibitem{XL2013}
X.~Huang and L.~Zhang, ``An svm ensemble approach combining spectral,
  structural, and semantic features for the classification of high-resolution
  remotely sensed imagery,'' \emph{IEEE Trans. Geosci. Remote Sens.}, vol.~51,
  no.~1, pp. 257--272, 2013.

\bibitem{PR2010}
P.~Zhong and R.~Wang, ``Learning conditional random fields for classification
  of hyperspectral images,'' \emph{IEEE Trans. Image Process.}, vol.~19, no.~7,
  pp. 1890--1907, 2010.

\bibitem{YA2013}
Y.~Bengio, A.~Courville, and P.~Vincent, ``Representation learning: A review
  and new perspectives,'' \emph{IEEE Trans. Pattern Anal. Mach. Intell.},
  vol.~35, no.~8, pp. 1798--1828, 2013.

\bibitem{DS2014}
D.~Lunga, S.~Prasad, M.~M. Crawford, and O.~Ersoy, ``Manifold-learning-based
  feature extraction for classification of hyperspectral data: A review of
  advances in manifold learning,'' \emph{IEEE Signal Process. Mag.}, vol.~31,
  no.~1, pp. 55--66, 2014.

\bibitem{XB2013}
X.~Jia, B.-C. Kuo, and M.~M. Crawford, ``Feature mining for hyperspectral image
  classification,'' \emph{Proc. IEEE}, vol. 101, no.~3, pp. 676--697, 2013.

\bibitem{LL2013a}
L.~Zhang, L.~Zhang, D.~Tao, and X.~Huang, ``Tensor discriminative locality
  alignment for hyperspectral image spectral-spatial feature extraction,''
  \emph{IEEE Trans. Geosci. Remote Sens.}, vol.~51, no.~1, pp. 242--256, 2013.

\bibitem{JX2015}
J.~Li, X.~Huang, P.~Gamba, J.~M. Bioucas-Dias, L.~Zhang, J.~A. Benediktsson,
  and A.~Plaza, ``Multiple feature learning for hyperspectral image
  classification,'' \emph{IEEE Trans. Geosci. Remote Sens.}, vol.~53, no.~3,
  pp. 1592--1606, 2015.

\bibitem{LQ2015}
L.~Zhang, Q.~Zhang, L.~Zhang, D.~Tao, X.~Huang, and B.~Du, ``Ensemble manifold
  regularized sparse low-rank approximation for multiview feature embedding,''
  \emph{Pattern Recognit.}, vol.~48, no.~10, pp. 3102--3112, 2015.

\bibitem{JH2015}
J.~Li, H.~Zhang, and L.~Zhang, ``Efficient superpixel-level multitask joint
  sparse representation for hyperspectral image classification,'' \emph{IEEE
  Trans. Geosci. Remote Sens.}, vol.~53, no.~10, pp. 5338--5351, 2015.

\bibitem{LX2016}
L.~Zhang, X.~Zhu, L.~Zhang, and B.~Du, ``Multidomain subspace classification
  for hyperspectral images,'' \emph{IEEE Trans. Geosci. Remote Sens.}, vol.~54,
  no.~10, pp. 6138--6150, 2016.

\bibitem{G1968}
G.~F. Hughes, ``On the mean accuracy of statistical pattern recognizers,''
  \emph{IEEE Trans. Inf. Theory}, vol.~14, no.~1, pp. 55--63, 1968.

\bibitem{YG2015}
Y.~Yuan, G.~Zhu, and Q.~Wang, ``Hyperspectral band selection by multitask
  sparsity pursuit,'' \emph{IEEE Trans. Geosci. Remote Sens.}, vol.~53, no.~2,
  pp. 631--644, 2015.

\bibitem{MJ2008}
M.~Fauvel, J.~A. Benediktsson, J.~Chanussot, and J.~R. Sveinsson, ``Spectral
  and spatial classification of hyperspectral data using svms and morphological
  profiles,'' \emph{IEEE Trans. Geosci. Remote Sens.}, vol.~46, no.~11, pp.
  3804--3814, 2008.

\bibitem{LL2013b}
L.~Zhang, L.~Zhang, D.~Tao, and X.~Huang, ``A modified stochastic neighbor
  embedding for multi-feature dimension reduction of remote sensing images,''
  \emph{ISPRS J. Photogramm.}, vol.~83, pp. 30--39, 2013.

\bibitem{TD2009}
T.~Zhang, D.~Tao, X.~Li, and J.~Yang, ``Patch alignment for dimensionality
  reduction,'' \emph{IEEE Trans. Knowl. Data Eng.}, vol.~21, no.~9, pp.
  1299--1313, 2009.

\bibitem{FH2010}
F.~Nie, H.~Huang, X.~Cai, and C.~H. Ding, ``Efficient and robust feature
  selection via joint l2,1-norms minimization,'' in \emph{Proc. NIPS}, 2010,
  pp. 1813--1821.

\bibitem{CD2015}
C.~Xu, D.~Tao, and C.~Xu, ``Multi-view intact space learning,'' \emph{IEEE
  Trans. Pattern Anal. Mach. Intell.}, vol.~37, no.~12, pp. 2531--2544, 2015.

\bibitem{DX2011}
D.~Cai, X.~He, J.~Han, and T.~S. Huang, ``Graph regularized nonnegative matrix
  factorization for data representation,'' \emph{IEEE Trans. Pattern Anal.
  Mach. Intell.}, vol.~33, no.~8, pp. 1548--1560, 2011.

\bibitem{JD2012}
J.~Yu, D.~Liu, D.~Tao, and H.~S. Seah, ``On combining multiple features for
  cartoon character retrieval and clip synthesis,'' \emph{IEEE Trans. Syst. Man
  Cybern. Part B Cybern.}, vol.~42, no.~5, pp. 1413--1427, 2012.

\bibitem{YH2014}
Y.~Y. Tang, H.~Yuan, and L.~Li, ``Manifold-based sparse representation for
  hyperspectral image classification,'' \emph{IEEE Trans. Geosci. Remote
  Sens.}, vol.~52, no.~12, pp. 7606--7618, 2014.

\bibitem{XP2004}
X.~He and P.~Niyogi, ``Locality preserving projections,'' in \emph{Proc. NIPS},
  vol.~16, 2004, pp. 153--160.

\bibitem{TD2010}
T.~Xia, D.~Tao, T.~Mei, and Y.~Zhang, ``Multiview spectral embedding,''
  \emph{IEEE Trans. Syst. Man Cybern. Part B Cybern.}, vol.~60, no.~6, pp.
  1438--1446, 2010.

\bibitem{YD2015}
Y.~Luo, D.~Tao, K.~Ramamohanarao, C.~Xu, and Y.~Wen, ``Tensor canonical
  correlation analysis for multi-view dimension reduction,'' \emph{IEEE Trans.
  Knowl. Data Eng.}, vol.~27, no.~11, pp. 3111--3124, 2015.

\bibitem{CD2014}
C.~Xu, D.~Tao, and C.~Xu, ``Large-margin multi-view information bottleneck,''
  \emph{IEEE Trans. Pattern Anal. Mach. Intell.}, vol.~36, no.~8, pp.
  1559--1572, 2014.

\bibitem{XY2013}
X.~Zhai, Y.~Peng, and J.~Xiao, ``Heterogeneous metric learning with joint graph
  regularization for cross-media retrieval,'' in \emph{Proc. AAAI}, 2013, pp.
  1198--1204.

\bibitem{HL2016}
H.~Shen, L.~Peng, L.~Yue, Q.~Yuan, and L.~Zhang, ``Adaptive norm selection for
  regularized image restoration and super-resolution,'' \emph{IEEE Trans.
  Cybern.}, vol.~46, no.~6, pp. 1388--1399, 2016.

\bibitem{CC2011}
C.-C. Chang and C.-J. Lin, ``Libsvm: A library for support vector machines,''
  \emph{ACM Trans. Intell. Syst. Technol.}, vol.~2, no.~3, p.~27, 2011.

\bibitem{LL2012}
L.~Zhang, L.~Zhang, D.~Tao, and X.~Huang, ``On combining multiple features for
  hyperspectral remote sensing image classification,'' \emph{IEEE Trans.
  Geosci. Remote Sens.}, vol.~50, no.~3, pp. 879--893, 2012.

\bibitem{JJ2005}
J.~A. Benediktsson, J.~A. Palmason, and J.~R. Sveinsson, ``Classification of
  hyperspectral data from urban areas based on extended morphological
  profiles,'' \emph{IEEE Trans. Geosci. Remote Sens.}, vol.~43, no.~3, pp.
  480--491, 2005.

\bibitem{HT2006}
H.~Zou, T.~Hastie, and R.~Tibshirani, ``Sparse principal component analysis,''
  \emph{J. Comput. Graphical Statist.}, vol.~15, no.~2, pp. 265--286, 2006.

\bibitem{LT2011}
L.~Clemmensena, T.~Hastieb, D.~Wittenc, and Bjarne, ``Sparse discriminant
  analysis,'' \emph{Technometrics}, vol.~53, no.~4, 2011.

\bibitem{JP2010}
J.-M. Yang, P.-T. Yu, and B.-C. Kuo, ``A nonparametric feature extraction and
  its application to nearest neighbor classification for hyperspectral image
  data,'' \emph{IEEE Trans. Geosci. Remote Sens.}, vol.~48, no.~3, pp.
  1279--1293, 2010.

\bibitem{HB2010}
H.-Y. Huang and B.-C. Kuo, ``Double nearest proportion feature extraction for
  hyperspectral-image classification,'' \emph{IEEE Trans. Geosci. Remote
  Sens.}, vol.~48, no.~11, pp. 4034--4046, 2010.

\bibitem{HW2014}
H.~Zhang, W.~He, L.~Zhang, H.~Shen, and Q.~Yuan, ``Hyperspectral image
  restoration using low-rank matrix recovery,'' \emph{IEEE Trans. Geosci.
  Remote Sens.}, vol.~52, no.~8, pp. 4729--4743, 2014.

\end{thebibliography}

\end{document}